\documentclass[review]{elsarticle}

\usepackage{lineno,hyperref}
\usepackage{url}
\modulolinenumbers[5]

\journal{Journal of Information Processing and Management}

\usepackage[T1]{fontenc}
\usepackage{soul}
\usepackage{tabularx}
\usepackage{multirow}
\usepackage[normalem]{ulem}
\usepackage{xcolor}
\definecolor{light-gray}{gray}{0.95}
\usepackage{enumitem}
\usepackage{graphicx}
\usepackage{amsmath}
\usepackage{amsthm}
\usepackage{amssymb}
\usepackage{stix}
\usepackage{booktabs}
\usepackage{array}
\usepackage{algorithm}
\usepackage[noend]{algpseudocode}
\usepackage{hhline}
\usepackage{wasysym}
\usepackage{arydshln}
\newcolumntype{M}[1]{>{\centering\arraybackslash}m{#1}}
\newcommand\mc[1]{\multicolumn{1}{l|}{#1}}

\biboptions{authoryear}

\begin{document}

\begin{frontmatter}

\title{Changing the Narrative Perspective:\\\textcolor{black}{From Deictic to Anaphoric Point of View}}
%\tnotetext[mytitlenote]{This is a footnote associated with the title.}


\author[mcaddress]{Mike Chen}
\ead{mc277509@ohio.edu}

\author[rbaddress]{Razvan Bunescu\corref{correspondingauthor}}
\cortext[correspondingauthor]{Corresponding author.}
\ead{rbunescu@uncc.edu}

\address[mcaddress]{School of Electrical Engineering and Computer Science, Ohio University}
\address[rbaddress]{Department of Computer Science, University of North Carolina at Charlotte}

\begin{abstract}
We introduce the task of changing the narrative point of view, where characters are assigned a narrative perspective that is different from the one originally used by the writer. The resulting shift in the narrative point of view alters the reading experience and can be used as a tool in fiction writing or to generate types of text ranging from educational to self-help and self-diagnosis. We introduce a benchmark dataset containing a wide range of types of narratives annotated with changes in point of view \textcolor{black}{from deictic (first or second person) to anaphoric (third person)} and describe a pipeline for processing raw text that relies on a neural architecture for mention selection. Evaluations on the new benchmark dataset show that the proposed architecture substantially outperforms the baselines by generating mentions that are less ambiguous and more natural.
\end{abstract}

\begin{keyword}
narrative perspective, narrative point of view
%\texttt{elsarticle.cls}\sep \LaTeX\sep Elsevier \sep template
%\MSC[2010] 00-01\sep  99-00
\end{keyword}

\end{frontmatter}

%\linenumbers

\section{Introduction and Motivation}
\label{sec:intro}

The narrative point of view (PoV), also called the narrative perspective, is the position from which the events in a story are observed and communicated \citep{kn:Willie01}. There are three types of narrative perspective, mirroring the use of personal pronouns: {\it first}, {\it second}, and {\it third} person. The narrative point of view influences the readers' perspective-taking and their involvement in the characters’ thoughts and feelings. In fiction, the most common narrative perspectives are the first and third person. The following sentence, taken from \citep{kn:Michael77}, shows a change of the PoV of the main character, a war correspondent in Vietnam, from the original $1^{st}$ person to $3^{rd}$ person:

\hangindent=0.4cm \emph{That night, probably sleeping, \{I\} $\rightarrow$ \{Michael\} heard the sound of automatic-weapons fire outside. \{I\} $\rightarrow$ \{He\} had no real sense of waking, only of suddenly seeing three cigarettes glowing in the dark.}

\noindent The $1^{st}$ person narration makes the reader feel like they are watching the story unfold through the eye of the journalist, who is the main protagonist. While the writer expresses his feelings about the events he goes through, the focus is on what he witnesses. Changing the PoV to $3^{rd}$ person, and referring to him as {\it Michael}, takes some of that focus away and puts it on the journalist.

The second person point of view is more common in poetry, how-to guides, technical
writing, and self-help texts, whereas in fiction the bulk of second person literature dates from only the past sixty years \citep{fludernik:aaa93}. In one common instantiation of this type of narrative perspective, the reader is incorporated as a character in the story and is addressed by the writer using second person pronouns. The following sentences, taken from \citep{kn:Melanie}, use the $2^{nd}$ person PoV to address the reader as a potential victim of narcissism:

\hangindent=0.4cm \emph{It doesn't matter how hard \{you\} try to do the right thing, the narcissist still finds fault with \{you\}. In stark contrast, \{you\} thought \{you\} could do no wrong with this person.}

\noindent While in general the $1^{st}$ person perspective leads to a more immersive reading experience than a $3^{rd}$ person perspective, the $2^{nd}$ person point of view used in the text above is even stronger in that it facilitates an identification of the reader with the protagonist. \cite{fludernik:aaa93} distinguishes between two major uses of the second person pronoun in narratives: the addressee {\it you} and the protagonist {\it you}, where the addressee can be further refined to the self-addressing {\it you} in interior monologue fiction. She argues that the initial ambiguity between the interpretation of {\it you} as the reader vs. a protagonist can develop into an increased empathy effect, achieving maximum identification in present tense texts. This can be used, for example, to manipulate readers into an empathetic identification with the protagonist from which it then proves difficult to withdraw when an important aspect of that character is finally revealed, a technique that forces the reader to become aware of their own prejudice and then proceed to either change (and thus stay consistent with the initial empathetic reading) or withdraw the empathy altogether.

The point of view is important in other modalities as well. In video games, the choice between the first vs. third person visual point of view was shown to moderate psychophysiological responses during play, with avatar choice producing a more pronounced effect in the third-person PoV (where the “camera” was located behind the avatar) \citep{lim:mp09}, while the first-person PoV was observed to heighten realism by providing a sense of inhabiting the representational space. The viewpoint use is much more dynamic in film, where it is very common for it to shift from first to third person, as for example in a shot–reverse shot sequence, where we are initially placed to observe events from the position of the film’s protagonist, only to be looking back at the protagonist from the viewpoint of another character moments later \citep{black:gc17}.

In the philosophy of mind, the three perspectives are considered to reflect different modes of epistemic access: the first person perspective is {\it subjective} because it is directed at the epistemic subject's experiences; the third person perspective is {\it objective} because it provides access to objective knowledge of other entities; the second person perspective is {\it intersubjective} because "it is an epistemic relation between an epistemic subject and another sentient being’s mental states" \citep{pauen:inquiry12}. Whereas the second person perspective has been relatively less studied \citep{bohman:chapter18}, the epistemic differences between the first and third person perspective are well established (knowing ourselves vs. knowing the world around us), with \citet{choifer:pp18} proposing that both belong to a {\it reflective} mode of experience, as opposed to the 'what-it-is-like' {\it non-reflective}, non-epistemic access to one's conscious experience. 

% you stand in for a character with whom the actual reader is led to empathise

% The  manipulation of narrative perspective can be a valuable tool in fiction. Of a particular interest in naratology and the philosophy of mind recently is the use of the second person perspective,

In narratives, the choice of PoV has been shown in multiple empirical studies to have a substantial effect on modulating the reader's engagement, perspective-taking, and overall their immersion in the events and the mental states described in a story. \cite{Lissa:16} studied how narrative perspective affects readers' responses to a complex, and potentially unreliable character. Their experimental results showed that the manipulation of narrative perspective influences readers' trust for the character. \cite{Salem:17} ran questionnaire studies to investigate the effect of different textual modes of narration on reading experience. Their work showed that first person narration and psycho-narration (a third person perspective of a character's consciousness) increased the tendency to take over the perspective of the protagonist.  \cite{Hartung:16} investigated the effect of personal pronouns on discourse comprehension, with results showing that first person pronouns lead to higher immersion. Using simple event sentences, \cite{Brunye:09} found that third person pronouns facilitate the construction of a mental model from an observer's perspective, whereas second person pronouns have the strongest effect in making the readers feel they embody an actor's perspective.

\begin{table}[t]
	\small	
	\begin{tabularx}{\textwidth}{|c|X|}
%		\hline%
%		PoV & Sample Text \\
%		\hline
		\hline
		\multicolumn{2}{|c|}{Autobiography $\leftrightarrow$ Biography}\\
		\hline
		$1^{st}$ & $\smblkdiamond$ When the youngest of \textit{my} sisters saw \colorbox{light-gray}{him} in his bloodied condition, \textit{I} clearly \dashuline{recall} her bursting out "Let me die in his place!". \citep{kn:Kurosawa83}\\
		%\hline%
		$3^{rd}$ & $\smblkdiamond$ When the youngest of \textit{Akira’s} sisters saw \colorbox{light-gray}{her brother} in his bloodied condition, \textit{Akira} clearly \dashuline{recalls} her bursting out "Let me die in his place!".\\
		\hline%
		\hline
		\multicolumn{2}{|c|}{Educational $\leftrightarrow$ Self-diagnosis}\\
		\hline%
		$3^{rd}$ & $\smblkdiamond$ \textit{Persons afflicted with the classic form of the illness} have periods of severe depression as well as periods of mania. \citep{kn:Mondimore06}\\
		%\hline%
		$2^{nd}$ & $\smblkdiamond$ \textit{You} have periods of severe depression as well as periods of mania.\\
		\hline%
		\hline%
%		$1^{st}$ & \textit{I} \dashuline{think} \textit{my} father's attitude toward films reinforced \textit{my} own inclinations and encouraged \textit{me} to become what \textit{I} \dashuline{am} today.\\
%		\hline%
%		$3^{rd}$ & \textit{Akira} \dashuline{thinks} \textit{his} father's attitude toward films reinforced \textit{his} own inclinations and encouraged \textit{him} to become what \textit{he} \dashuline{is} today.\\
%		\hline%
%		\hline%
		\multicolumn{2}{|c|}{Abstract $\leftrightarrow$ Related work}\\
		\hline
		$1^{st}$ & $\smblkdiamond$ \textit{We} show that the bound is tight, and \textit{we} develop an algorithm that computes the optimal schedule. \citep{kn:Gast13}\\
		%\hline%
		$3^{rd}$ & $\smblkdiamond$ \textit{The authors} show that the bound is tight, and \textit{they} develop an algorithm that computes the optimal schedule.\\
		\hline%
	\end{tabularx}
	\caption{Pairs of sentences written from different narrative perspectives. Verbs whose conjugation is changed due to number and person agreement are shown with \protect\dashuline{dotted underlines}.}
	\label{table:sample-texts-of-different-perspectives}
\end{table}

\subsection{On the utility of changing the narrative perspective}

Given the substantial effect that PoV has on the readers' perspective-taking, the manipulation  of  narrative  perspectives can be a valuable tool in fiction writing. Furthermore, in non-fiction, articles written from different perspectives can serve different purposes, some of which are illustrated in  Tables~\ref{table:sample-texts-of-different-perspectives} and~\ref{table:sample-texts-relational}. The first example in Table~\ref{table:sample-texts-of-different-perspectives} is extracted from an autobiography. By converting from $1^{st}$ to $3^{rd}$ person PoV, the text could be used for a biography or for an encyclopedia, depending on the importance and the level of detail of the factual information conveyed.
The second example is extracted from an educational text describing symptoms of a bipolar personality disorder. By converting the main character's PoV from $3^{rd}$ to $2^{nd}$ person, the text becomes more suitable for readers who are interested in interpreting the symptoms in the context of their own behavior, i.e. self-diagnosis. The third example shows a sample from the abstract section of a paper, where it is common for authors to use the $1^{st}$ person PoV. By converting it to $3^{rd}$ person, the transformed text becomes suitable for the related work section of another paper. Note that the direction of the PoV change can be reversed, transforming the second text into the first text in each of these example pairs.

\begin{table}[t]
	\small	
	\begin{tabularx}{\textwidth}{|X|}
		\hline
		\multicolumn{1}{|c|}{$E_1$ in $2^{nd}$, $E_2$ in $3^{rd}$ $\longleftrightarrow$ $E_1$ in $3^{rd}$, $E_2$ in $2^{nd}$}\\
		\hline%
		\hline%
		% Here are 5 signs of a trustworthy lawyer.
		$\smblkdiamond$ Most cases aren't slam-dunks, and it is important that {\it $[$your lawyer$]_2$} \dashuline{doesn't} make promises regarding the outcome of {\it $[$your$]_1$} case and should not be overconfident no matter how seasoned {\it $[$he or she$]_2$} \dashuline{is}. \citep{website:lawyer}.\\
		%\hline%
		% Here are 5 signs of a trustworthy lawyer.
		$\smblkdiamond$ Most cases aren't slam-dunks, and it is important that \textit{$[$you$]_2$} \dashuline{don't} make promises regarding the outcome of \textit{$[$your client's$]_1$} case and should not be overconfident no matter how seasoned \textit{$[$you$]_1$} \dashuline{are}.\\
		\hline%
		\hline%
		% Here are some of the essential elements of a healthy doctor-patient relationship.
		$\smblkdiamond$ Communication begins from the moment \textit{$[$you$]_1$} first \dashuline{meet} \textit{$[$your doctor$]_2$}. \dashuline{Does} \textit{$[$he or she$]_2$} greet \textit{$[$you$]_1$} warmly? \citep{website:doctor}\\
		% Here are some of the essential elements of a healthy doctor-patient relationship.
		$\smblkdiamond$ Communication begins from the moment \textit{$[$your patient$]_1$} first \dashuline{meets} \textit{$[$you$]_2$}. \dashuline{Do} \textit{$[$you$]_2$} greet \textit{$[$her$]_1$} warmly?\\
		\hline%
		\hline%
		$\smblkdiamond$ \textit{$[$The toxic mother$]_2$} \dashuline{is} also likely to spot the speck in an otherwise perfect offering, and \textit{$[$her$]_2$} perfectionism will cause \textit{$[$you$]_1$} to never quite to feel good enough, no matter what \textit{$[$you$]_1$} \dashuline{do}. \citep{website:mother}\\
		$\smblkdiamond$ \textit{$[$You$]_2$} \dashuline{are} also likely to spot the speck in an otherwise perfect offering, and \textit{$[$your$]_2$} perfectionism will cause \textit{$[$your child$]_1$} to never quite to feel good enough, no matter what \textit{$[$she$]_1$} \dashuline{does}.\\
		\hline
	\end{tabularx}
	\caption{Changing from $3^{rd}$ to $2^{nd}$ person PoV in texts already containing a $2^{nd}$ person PoV. Verbs whose conjugation is changed due to number and person agreement are shown with \protect\dashuline{dotted underlines}.}
	\label{table:sample-texts-relational}
\end{table}

The narrative examples from Table~\ref{table:sample-texts-relational} have two major characters, or roles, $E_1$ and $E_2$, whose thoughts and behaviors are originally described using the second person point of view for $E_1$ and the third person point of view for $E_2$. The two roles are in a particular relationship, e.g. client vs. lawyer, patient vs. doctor, or child vs. toxic mother. In each example pair the narrative perspectives are switched such that role $E_2$ is assigned the second person PoV, while role $E_1$ receives the third person PoV, effectively changing the target audience of the text. Thus, the first text, which was originally intended to help clients recognize trustworthy lawyers, becomes more suitable for lawyers interested in earning the trust of their clients.

The adaptation of existing stories to satisfy the requirements of a new medium provides an additional application area. A large number of games are played from a second-person viewpoint \citep{harrigan:book07}. However, many of these games are based on previously written stories where the protagonist uses the third-person PoV. Changing the narrative perspective could assist in the adaptation of the existing story in order to generate the game's introduction narrative or script. 

In light of the above, in this paper we propose the task of {\it changing the narrative perspectives} of a text. We believe that developing a system that automatically changes the points of view would \textcolor{black}{enable new types of narrative generation tasks. For example, a literature review generation system would change the narrative perspective from $1^{st}$ to $3^{rd}$ person in the abstract and conclusion sections that are extracted automatically from papers on a particular topic, possibly in tandem with a summarization or paraphrasing step. Another potential application would be the automatic generation or updating of Wikipedia articles using information extracted from biographies, where the $1^{st}$ person PoV would need to be changed to the more neutral $3^{rd}$ to make it suitable for an encyclopedic text. Although excluded from the task of {\it narrative} perspective change considered in this paper, the automatic processing of dialogues presents additional opportunities for a PoV change system. In one application, the system would be tasked with translating dialogues, where the $1^{st}$ person PoV is pervasive, into an indirect speech version requiring a $3^{rd}$ person PoV. This could be useful, for example, in generating meeting notes or summaries of meetings where two or more people interact directly through dialogue.}

\textcolor{black}{\subsection{Outline and main contributions}}

\textcolor{black}{In the next section, we position the new task of changing the narrative perspective in the wider context of related work on text generation and narrative processing (Section~\ref{sec:related-work}). The following sections in the paper are then focused on its main contributions:}
\begin{enumerate}%[topsep=0cm,itemsep=0cm,parsep=0cm]
	\item Characterizing the new task of changing the narrative perspectives of a text (Section~\ref{sec:definition}).
	\item Designing a neural architecture for mention selection (Section~\ref{sec:neural}) that lies at the core of an end-to-end system \textcolor{black}{for deictic to anaphoric PoV change}, where the text is first pre-processed for mention detection, coreference resolution, and verb conjugation (Section~\ref{sec:system}).
	\item Showing how coreference annotations can be used to create training examples for the mention selection task (Section~\ref{sec:conll-data}).
	\item Introducing a new benchmark dataset comprised of a diverse set of types of narratives that are manually annotated with \textcolor{black}{ deictic to anaphoric} PoV changes (Section~\ref{sec:pov-data}).
	\item Evaluations showing the proposed architecture outperforms decision tree-based approaches and other baselines, by virtue of generating mentions that lead to text that is more fluent and less ambiguous (Section~\ref{sec:evaluation}).
\end{enumerate}
The paper ends with ideas for future work and concluding remarks (Section~\ref{sec:conclusion}).

\section{Related Work}
\label{sec:related-work}

\citet{kn:Wiebe94} addressed the problem of tracking the psychological point of view in third-person fictional narrative text, which is about recognizing sentences that convey the thoughts, perceptions, and inner states of characters and mapping these psychological perspectives to the corresponding character in the story. \citet{kn:Mani13} provided a systematic overview of computational methods applied to the processing of core narratological concepts, such as narrative perspective, narrative threads, and narrative time. An annotation scheme, NarrativeML, was developed to facilitate the annotation of these concepts. More recently, \citet{kn:eisenberg2016} trained SVM models to distinguish between first-person and third-person point of view in a narrative, as well as for determining whether the narrator is involved in the story, i.e. diegesis.
		
Most related to the PoV change is the task of generating referential expressions in context (GREC) \citep{belz:chapter10}. In one version of this task, Wikipedia articles provide the context, referential expressions to the main subject of the article are erased, and lists of possible referring expressions are provided. The task then is to select referring expressions to insert into the gaps such that the resulting text is fluent and coherent. In all GREC tasks, the PoV of the focus entity is always $3^{rd}$ person, and the context, including mentions of other entities, does not have to change. The task of changing of narrative perspective, however, is more complicated in that the change from a deictic PoV ($1^{st}$ or $2^{nd}$) to an anaphoric PoV ($3^{rd}$) may require changing the context. As explained in Section~\ref{sec:interactions}, deictic pronouns are agnostic of centering constraints, therefore changing them to anaphoric pronouns may lead to referential ambiguity, unless mentions of other entities are changed too. NeuralREG \citep{ferreira:acl18} is another relevant system, where the task is to generate referential expression in a fixed context, a delexicalized version of the WebNLG corpus \citep{gardent:acl17}. In most instances, the context is composed of one or two sentences that describe in natural language the knowledge encoded in a set of RDF triples extracted from DBPedia. In contrast, the contexts in the PoV task are much longer and need to change to accommodate replacing deictic pronouns with anaphoric pronouns. Furthermore, NeuralREG uses a seq2seq decoder to generate referential expressions for entities that appear in both the training and test data. In contrast, by the definition of the PoV change task, the approach presented in this paper works with new entities at test time.
		
Also related to the PoV Change task is style transfer, where a text is transformed automatically from one particular style into another. The transferred style can refer to: formality \citep{kn:Heylighen99formalityof,kn:rao-tetreault-2018-dear}, modern English to Shakespeare English \citep{kn:xu-etal-2012-paraphrasing,kn:JhamtaniGHN17}, sentiment \citep{kn:Fu17,kn:Shen17}, or gender and political slant \citep{kn:prabhumoye-etal-2018-style}.

Approaches to the PoV change task may benefit from techniques that are used in the general task of entity generation. \citet{kn:ji-etal-2017-dynamic} presented a new type of language model, EntityNLM, which can 
% explicitly model entities. 
model an arbitrary number of entities in context while generating each entity mention at an arbitrary length. 
%Their system can be used for language modeling, coreference resolution and entity prediction.
\citet{kn:clark-etal-2018-neural} introduced an approach to neural text generation that explicitly represented entities mentioned in the text, which was shown to benefit mention generation.

\textcolor{black}{Overall, the new task of changing narrative perspectives can be seen as part of a larger interdisciplinary area concerned with computational approaches to storytelling and narrative processing. A number of diverse tasks being addressed in this area and related fields have been explored in the Storytelling Workshop series \citep{ws-2018-storytelling}, as well as within this journal in its Special Issue on Narrative Extraction from Texts \citep{jorge_information_2019}. Finally, very recent developments in the field are to be presented in the upcoming Fourth International Workshop on Narrative Extraction from Texts (Text2Story) \citep{ws-2021-storytelling}.}

\section{Task Definition and Important Aspects}
\label{sec:definition}

We consider as input a text document \textcolor{black}{in a prototypical PoV setting} where at most one discourse entity has the $1^{st}$ person PoV, at most one entity has the $2^{nd}$ person PoV, and any number of entities have the $3^{rd}$ person PoV\footnote{\textcolor{black}{When counting the number of entities with a $1^{st}$ or $2^{nd}$ person PoV, we do not consider entity mentions within dialogues or direct speech, which are normally unaffected by the PoV change (as shown in some of the examples in this paper).}}.
Given a focus entity ${E}$ that has PoV $\mathit{p}$, the task is to change the text such that ${E}$ is assigned a different PoV $\mathit{q}$. Changing the PoV of the focus entity may also require changing mentions of other entities, as needed to eliminate ambiguity or to preserve the naturalness of the original text. We will use the term {\it confounding entities} to refer to entities other than the focus entity that are affected by the PoV change. In the first example shown in Table~\ref{table:sample-texts-of-different-perspectives}, the focus entity is ``Akira Kurosawa". If the pronoun ``him" were left unchanged, its referent would be ambiguous between {\it Akira} and the confounding entity {\it his brother}. The approach that will be described in Section~\ref{sec:system} however has the capability to convert ``him" to ``her brother", eliminating part of the ambiguity.

\textcolor{black}{Overall, the task of changing the PoV of a focus entity can be subcategorized into one of the following 3 categories:
\begin{itemize}
    \item {\bf Deictic to Deictic (D2D)}: PoV change from $1^{st}$ to $2^{nd}$ or from $2^{nd}$ to $1^{st}$.
    \item {\bf Anaphoric to Deictic (A2D)}: PoV change from $3^{rd}$ to $1^{st}$ or from $3^{rd}$ to $2^{nd}$.
    \item {\bf Deictic to Anaphoric (D2A)}: PoV change from $1^{st}$ to $3^{rd}$ or from $2^{nd}$ to $3^{rd}$.
\end{itemize}
In this paper, we focus on the deictic to anaphoric D2A case for the reason that it is often also a required step in the A2D and D2D transformations. As will be shown in the examples below, A2D (and similarly D2D) changes can be done in three steps:
\begin{enumerate}
    \item Substitution of a mention string with a deictic pronoun suitable for its case (e.g accusative) and number (e.g. singular);
    \item Potential change in conjugation of a verb modifying the original mention string;
    \item Deictic to anaphoric changes for confounding entities.
\end{enumerate}}
For example, given the coreference chains, changing an entity's PoV to the $2^{nd}$ person starts by replacing all mentions in its chain with the appropriate $2^{nd}$ person pronouns \textcolor{black}{(step 1)}. However, the discourse may have already referred to another (confounding) entity using the $2^{nd}$ person, in which case that entity's PoV needs to be changed to $3^{rd}$ person to avoid ambiguity \textcolor{black}{(step 3)}. In the examples from Table~\ref{table:sample-texts-relational}, in order to change the entity $E_2$ to use the $2^{nd}$ person PoV, entity $E_1$, which was already using the $2^{nd}$ person PoV, has to be changed to $3^{rd}$ person PoV. Overall, changing the PoV may require, directly or indirectly, a D2A change from the $2^{nd}$ person to the $3^{rd}$ person. The same reasoning applies when the target PoV is $1^{st}$ person. \textcolor{black}{In the examples below, the A2D task is to change the entity {\it Cliff} from a $3^{rd}$ to a $1^{st}$ person PoV. In the first example, the substitution is straightforward: all the corresponding entity mentions, which are in the nominative case, are changed to the pronoun {\it I}:
\begin{itemize}
    \item {\it $[$Cliff$]_1 \rightarrow [${\bf I}$]_1$ laughed. $[$He$]_1 \rightarrow [${\bf I}$]_1$ said, "I'm thinking of a cartoon I saw."}
\end{itemize}
In the second example, there is another (confounding) entity {\it Bill} that is already using the $1^{st}$ person PoV. This will need to be changed to use a different PoV (by default $3^{rd}$ person), to allow {\it Cliff} to take over the $1^{st}$ person PoV. This example also shows the need to change the conjugation of a verb to agree in person with the new PoV:
\begin{itemize}
    \item {\it $[$I$]_2 \rightarrow [${\bf Bill}$]_2$ asked something, then $[$he$]_1 \rightarrow [${\bf I}$]_1$ said something, $[$I$]_2 \rightarrow [${\bf Bill}$]_2$ asked some more and $[$he$]_1 \rightarrow [${\bf I}$]_1$ explained. The details don't matter, but as $[$I$]_2 \rightarrow [${\bf Bill}$]_2$ said, $[$he$]_1 \rightarrow [${\bf I}$]_1$ \dashuline{is $\rightarrow$ {\bf am}} the mathematician of the combination.}
\end{itemize}
The third example shows the need to choose a pronoun replacement appropriate for the accusative case:
\begin{itemize}
    \item {\it $[$I$]_2 \rightarrow [${\bf He}$]_2$ had been on the phone talking to $[$Cliff$]_1 \rightarrow [${\bf me}$]_1$ in the lab.}
\end{itemize}
Note that the agreement in case and number (step 1) and the potential change in verb conjugation (step 2) required for A2D and D2D transformations are also necessary for a successful implementation of the D2A transformation. Therefore, given the need of the D2A transformation (as step 3) in A2D and D2D, we believe that successfully addressing D2A will also provide the solution for a significant number, if not the majority, of A2D and D2D transformations. This is not to say, however, that there are no situations in which A2D or D2D would require a more specialized approach: consider the example below, where the change from $3^{rd}$ to $1^{st}$ person PoV results in the somewhat awkward phrase {\it "a German in my mid sixties"}:
\begin{itemize}
    \item {\it $[$This fellow$]_1 \rightarrow [${\bf I}$]_1$ was the son of a Nazi -- a German in $[$his$]_1 \rightarrow [${\bf my}$]_1$ mid sixties whose father had served in the SS during World War II.}
\end{itemize}
The example above further illustrates the complexity of the task of changing the narrative perspective. Ideally, a complete solution would have the capability to also transform the text between the mentions whenever that can preserve the naturalness and fluency of the original text. As a first step in solving the PoV change task, in this paper we consider a formulation where the change in narrative perspective is done by changing only mentions of focus entities and their confounding entities, leaving the more general transformation of text as a direction for future work.
}
%{\it $[$I$]_2 \rightarrow [${\bf He}$]_2$ was scribbling down some of what $[$Cliff$]_1 \rightarrow [${\bf I}$]_1$ had just told $[$me$]_2 \rightarrow [${\bf him}$]_2$ while $[$I$]_2 \rightarrow [${\bf he}$]_2$  watched.} 

% This is also the most complicated PoV change, as such it is the focus of this paper. 
The deictic to anaphoric PoV change that is the focus of this paper is a non-trivial task that requires selecting a third person pronoun or noun phrase to replace mentions of the focus entity $E$  such that the resulting text satisfies two major criteria:
\begin{itemize}%[topsep=0cm,itemsep=0cm,parsep=0cm]
	\item[$\circ$] {\bf Manageable ambiguity}: The selected strings lead to non-ambiguous references, e.g. when a pronoun is generated, the reader can easily and correctly infer the entity it refers to.
	\item[$\circ$] {\bf Natural text}: The text sounds natural, which \textcolor{black}{generally means that it is grammatical and not repetitive in terms of the words used for mentions of the same entity}. %  which for example prohibits using the same noun phrase all the time.
\end{itemize}
\textcolor{black}{In Section~\ref{sec:human-evaluation} we provide examples illustrating the various levels of referential ambiguity and naturalness used for the human evaluations. These examples were included in the annotation guidelines in order to help the human annotators reliably assign referential ambiguity and naturalness scores.}

\subsection{Interactions with Other Aspects of Language}
\label{sec:interactions}

The task of changing the PoV can be further complicated by {\it quotations}, {\it dialogues}, {\it generic pronouns}, and the use of {\it narrator's perspective}. 

The text inside quotations and dialogues is usually kept unchanged when converting the PoV, however identifying dialogues and quotations can be difficult \cite{pareti-etal-2013-automatically}, especially when no formatting, e.g. quotation marks, is available. Likewise, pronouns used in a generic sense are kept unchanged when converting the PoV, however identifying generic expressions is still a very difficult NLP task -- \citet{friedrich-pinkal-2015-discourse}, for instance, achieve a state-of-the-art of less than 60\% F1 score. The PoV also needs to stay unchanged for the narrator's perspective, as will be shown in the {\it Mother Night} \cite{kn:Vonnegut99} example from Section~\ref{sec:pov-data}. There, it is clear that the narration uses the voice of the author, and thus the PoV needs to stay in the $1^{st}$ person.

\citet{smith_2003} catalogued five discourse modes: {\it Narrative, Description, Report, Information} and {\it Argument}. The events in Narratives are temporally related according to narrative time, whereas in Descriptions the text progression is spatial rather than temporal. In Reports, the time of the situations expressed is evaluated against speech time, whereas the Information and Argument modes are atemporal. The dataset created for the task of PoV change is mainly based on texts in the Narrative mode, where the alternation between different timelines may introduce additional difficulties -- such an example will be shown in Section ~\ref{sec:pov-data}, \textcolor{black}{where the narrative alternates between the past (the time of the events in the story, where the PoV changes) and the present (the narrator's time, where the author's voice is used extensively and therefore the original PoV needs to be preserved).}

In this paper we focus on changing the PoV in English. While many aspects of the proposed approach can transfer to other languages, changes may be required depending on properties of the actual language. For example, in pro-drop languages, pronouns in subject position are allowed to be dropped in certain contexts. Therefore, in addition to overt pronouns and noun phrases, the mention selection step would also need to consider zero pronouns. There is usually a flexible word order in languages that have extensive case systems and use a high degree of morphological marking to disambiguate the semantic roles of the arguments. In such languages, word order changes may be required when changing the PoV. For example, when replacing a first person pronoun with a proper name, the case and morphological marking as well as position may differ from the original. Furthermore, coming up with unambiguous references for PoV changes will be more difficult in languages that do not make gender distinctions.

Similar to the task of generating referential expression in context (GREC) \citep{belz:chapter10}, the task of changing the PoV requires selecting the form of referential expression: pronoun vs. description vs. proper name. The Centering Theory \citep{grosz-etal-1995-centering} relates focus of attention, choice of referring expression, and perceived coherence of utterances within a discourse segment. Accordingly, discourse entities that are {\it salient} in the current context are more likely to be realized in text using shorter expressions, such as pronouns. A similar conclusion is reached using other psycholinguistics models of reference, such as the Uniform Information Density (UID) hypothesis according to which speakers tend to convey information at a uniform rate \citep{kn:Jaeger10,kn:Jaeger16}. Applied to the choice of referring expressions, UID entails that a highly {\it predictable} referent should be encoded using a short code, e.g. a pronoun, while an unpredictable referent should be encoded using a longer form\footnote{For a more in-depth discussion of UID and factors that affect prediction, the reader is referred to \citep{modi:tacl17}}. However, unlike the GREC task where the PoV is always $3^{rd}$ person in a context that is given and does not need to change, in the PoV change task the context might have to change because of the switch between deictic pronouns ($1^{st}$ or $2^{nd}$) and anaphoric pronouns ($3^{rd}$). In the context of centering theory, \citet{poesi:cl04} found that $2^{nd}$ person pronouns behave differently from $3^{rd}$ person pronouns. One rule in centering says that if any of the forward-looking centers (CF) are realized as a pronoun in the current sentence, then its backward-looking center (CB) should be pronominalized too. However, the corpus-driven study of \cite{poesi:cl04} found that the rule does not apply to $2^{nd}$ person pronouns, confirming \cite{walker:centering93} who had suggested that they were not covered by centering. This is important for the task of PoV change, because a context that was structured for a $2^{nd}$ person PoV (which is agnostic of centering rules) might have to be altered when changed to a $3^{rd}$ person PoV to satisfy the centering rules and constraints, which specify conditions for the coherence of the resulting text. For example, in fixed-word-order languages such as English, syntactic position determines the partial ordering of the CF list, with subject ranked higher than objects, which in turn are ranked higher than other positions. When a new $3^{rd}$ person pronoun is introduced in a sentence, this might break the centering rule above, because syntactic positions positions are the same, and hence the ordering of the CF list is unchanged. Allowing other pronouns in the sentence to be replaced can alleviate this, as shown in the first example in Table~\ref{table:sample-texts-of-different-perspectives}, demonstrating the importance of developing a model that can change not only the focus entity mentions, but also realizations of other entities. In (relatively) free-word-order languages such as German, functional rather than syntactic criteria are important for the ranking of CF lists \citep{strube:cl99}, based on the distinction between hearer-old and hearer-new discourse entities \citep{prince:chapter81}. This seems to indicate that a PoV system designed for English might need to be adapted to work for free-word-order languages.

\section{The PoV Change System Pipeline}
\label{sec:system}

To change the narrative perspective from $1^{st}/2^{nd}$ to $3^{rd}$ PoV, the raw text is processed using the pipeline of 4 steps shown below. As a running example, we will use the text below, where the focus entity \textcolor{black}{(to be indexed with 2, and later named {\it Nick})} has the $1^{st}$ person PoV \textcolor{black}{(as expressed through the pronoun {\it I})} and the task is to convert it to a $3^{rd}$ person PoV:
\begin{itemize}
  \item[] {\it "Phil returns to Boston, to his job. I drive to the city every other week, to work a night or two at Pine Street, to see Emily."} \citep{kn:Flynn05}
\end{itemize}

\noindent {\bf Step 1}: Identify coreference chains corresponding to the focus and confounding entities.
\begin{itemize}[topsep=2pt,itemsep=2pt,partopsep=2pt, parsep=2pt, leftmargin=1.75\parindent]
  \item[1.1)] Perform coreference resolution \citep{kn:Lee17} and named entity recognition \citep{kn:Spacy} to extract all coreference chains referring to people. 

    {\it $[$Phil$]_1$ returns to Boston, to $[$his$]_1$ job. $[$I$]_2$ drive to the city every other week, to work a night or two at Pine Street, to see $[$Emily$]_3$.}

  \item[1.2)] Identify the coreference chain corresponding to the focus entity, based on the original PoV or the name of the focus entity.

    {\it $[$Phil$]_1$ returns to Boston, to $[$his$]_1$ job. $\mathbf{[I]_2}$ drive to the city every other week, to work a night or two at Pine Street, to see $[$Emily$]_3$.}

  \item[1.3)] Coreference chains that contain singular third person pronouns that match the gender of the focus entity will be considered confounding entities. \textcolor{black}{We assume the name and the gender of the focus entity are given as input. In general, the gender cannot be determined from the text only, as deictic pronouns can refer to any gender.}

    {\it $\mathbf{[Phil]_1}$ returns to Boston, to $\mathbf{[his]_1}$ job. $[$I$]_2$ drive to the city every other week, to work a night or two at Pine Street, to see $[$Emily$]_3$.}
\end{itemize}
In sub-step 1.3, an additional set of confounding entities will be created from coreference chains that contain plural pronouns that have the original PoV of the focus entity, e.g. if the original PoV is $1^{st}$ person, then a chain containing "we" will be considered as a confounding entity. Mentions inside quotations and dialogues are ignored, as well as pronouns that do not belong to any coreference chain (e.g. generic pronouns). First person pronouns that refer to the narrator are ignored as well by searching for patterns indicating a narrator's perspective, such as "I promise you", "I guarantee you", "I guess". A list of performative verbs \citep{kn:performative} are used to generate these patterns.

\noindent {\bf Step 2}: Change conjugation of verbs with focus entity mentions as subjects.
\begin{itemize}[topsep=2pt,itemsep=2pt,partopsep=2pt, parsep=2pt, leftmargin=1.75\parindent]
  \item[2.1)] Identify verbs whose subject is a focus entity mention and change their conjugation if necessary such that they agree in person with the new PoV, as shown underlined in Table~\ref{table:sample-texts-of-different-perspectives} and in bold in the example below.

    {\it $[$Phil$]_1$ returns to Boston, to $[$his$]_1$ job. $[$I$]_2$ \dashuline{\bf drive} $\rightarrow$ \dashuline{\bf drives} to the city every other week, to work a night or two at Pine Street, to see $[$Emily$]_3$.}

    To identify subject relations, we use a set of rules that leverage syntactic dependencies extracted by the dependency parser of \cite{DBLP:conf/iclr/DozatM17}.
  \item[2.2)] Verb conjugation is changed using the dictionary from \citep{manning-EtAl:2014:P14-5}, or a set of simple rules for out of vocabulary verbs, such as add 'es' to a verb ending in 's','sh','ch','x','z',or 'o'.
\end{itemize}

\noindent {\bf Step 3}: Create a set $S(E)$ of candidate mention strings for each focus or confounding entity $E$.
\begin{itemize}[topsep=2pt,itemsep=2pt,partopsep=2pt, parsep=2pt, leftmargin=1.75\parindent]
  \item[3.1)] For a focus entity $E$, we first add to $S(E)$ their full name, first and last name, whenever known. We then add pronouns that match their gender, i.e. either \{{\it she, her, herself}\} or \{{\it he, him, his, himself}\}. We further leverage the syntactic dependencies extracted with the parser used in step 2 to identify predicate nominals and appositives that involve the focus entity and add them to the candidate set, e.g. "the doctor" from "I was a doctor". Finally, we use a dictionary of relational nouns  \citep{newell-cheung-2018-constructing} to identify all possessive constructions where the focus entity modifies a relational noun, e.g. "my sister". Depending on the gender of the focus entity, we then add to its $S(E)$ the converse relational phrase, e.g. either "her brother" or "her sister". Furthermore, if the confounding entity "my sister" is coreferential with a name, e.g. "Mandy", we add the corresponding possessive construction too, e.g. "Mandy's brother" or "Mandy's sister".
  \item[3.2)] For a confounding entity $E$ that is $3^{rd}$ person singular, its candidate set $S(E)$ will contain the strings of all the mentions in its coreference chain. Confounding entities that have plural $1^{st}$ or $2^{nd}$ person PoV need to be changed to plural $3^{rd}$ person PoV, as such their $S(E)$ will contain all plural third person pronouns. Additionally, coordinated phrases that contain names will be transformed appropriately and added to the candidate set, e.g. an original occurrence of "Mandy and me" that corefers with "we" leads to "Mandy and him" or "Mandy and her" being added to $S(E)$, depending on the gender of the focus entity.
\end{itemize}
Setting $E_1$ to be the confounding entity {\it Phil} and $E_2$ to be the focus entity {\it Nick}, using the procedure above their candidate sets will be created as: $S(E_1)$ = \{{\it Phil}, {\it his}\} and $S(E_2)$ = \{{\it Nick Flynn, Nick, Flynn, Nick Flynn's, Nick's, Flynn's, he, him, his, himself, he himself, his son, her son, his brother, his grandson}\}.

\noindent {\bf Step 4}: Going from left to right, for each focus or confounding entity mention, select an appropriate string from $S(E)$, using the mention selection approach described in Section~\ref{sec:neural}.
\begin{itemize}[topsep=2pt,itemsep=2pt,partopsep=2pt, parsep=2pt, leftmargin=1.75\parindent]
  \item[4.1)] For each mention, we also use the original mention string to narrow down the choices in $S(E)$. For example, if the original mention string is a reflexive or possessive pronoun (e.g. "my" in "my sister"), then the choices of potential mention strings are constrained to reflexive pronouns or possessive pronouns and nouns, respectively (e.g. "Akira's" in "Akira's sister"). 
\end{itemize}
Using the same example, the final result will be:
\begin{itemize}
  \item[]  {\it $[$Phil$]_1$ returns to Boston, to $[$his$]_1$ job. $[$I$]_2$ $\rightarrow$ $[${\bf Nick}$]_2$ \dashuline{drive} $\rightarrow$ \dashuline{drives} to the city every other week, to work a night or two at Pine Street, to see $[$Emily$]_3$.}
\end{itemize}

\section{A Neural Mention Selection Model}
\label{sec:neural}

Given a mention position $t$ in the text for an entity $E$, the mention selection task is to select from the set $S(E)$ of possible strings for that entity the best string (noun phrase or pronoun) to use as a mention $M_t$ in that context. \textcolor{black}{Using the same example as in the previous section, where the focus entity $E$ is the one indexed with 2, and assuming that it was already mentioned 5 times before the sentences in this example, then the current mention position would be 6, as shown below:
\begin{itemize}
  \item[]  {\it $[$Phil$]_1$ returns to Boston, to $[$his$]_1$ job. $[M_6]_2$ \dashuline{drives} to the city every other week, to work a night or two at Pine Street, to see $[$Emily$]_3$.}
\end{itemize}
Its corresponding set of candidate string, narrowed down in sub-step 4.1 above, will be $S(E_2)$ = \{{\it Nick Flynn, Nick, Flynn, he, he himself, his son, her son, his brother, his grandson}\}.}

We train a scoring function $score(s|C)$ to capture how appropriate a string $s \in S(E)$ is to be used as a mention of an entity $E$ in a context C. 
Given $s_i, s_j \in S(E)$, we use $\langle s_i \succ s_j|C \rangle$ to denote that ${s_i}$ is more appropriate as a mention string than ${s_j}$ in the context $\mathit{C}$. Correspondingly, the ranking system is expected to rank ${s_i}$ higher than ${s_j}$, by computing a $score(s_i|C) > score(s_j|C)$. \textcolor{black}{In the example, this means that the system is expected to compute $score(\mbox{\it Nick}|C) > score(he|C)$ for mention $M_6$.}

In our experiments, we use the margin-based ranking loss shown below, where $\gamma$ is a tunable margin hyper-parameter:
\begin{equation}
\displaystyle
J(S(E)|C) = \sum_{s_i \succ s_j | C} \!\! max\{0, \gamma - score(s_i|C) + score(s_j|C)\} %\nonumber
\label{eq:ranking}
\end{equation}
At training time, we use the observed mention string $\hat{s} \in S(E)$ to create ranking pairs $\langle \hat{s} \succ s|C \rangle$, for all $s \in S(E), s \neq \hat{s}$, as training examples.

\begin{figure*}
\centering
	\includegraphics[width=\textwidth]{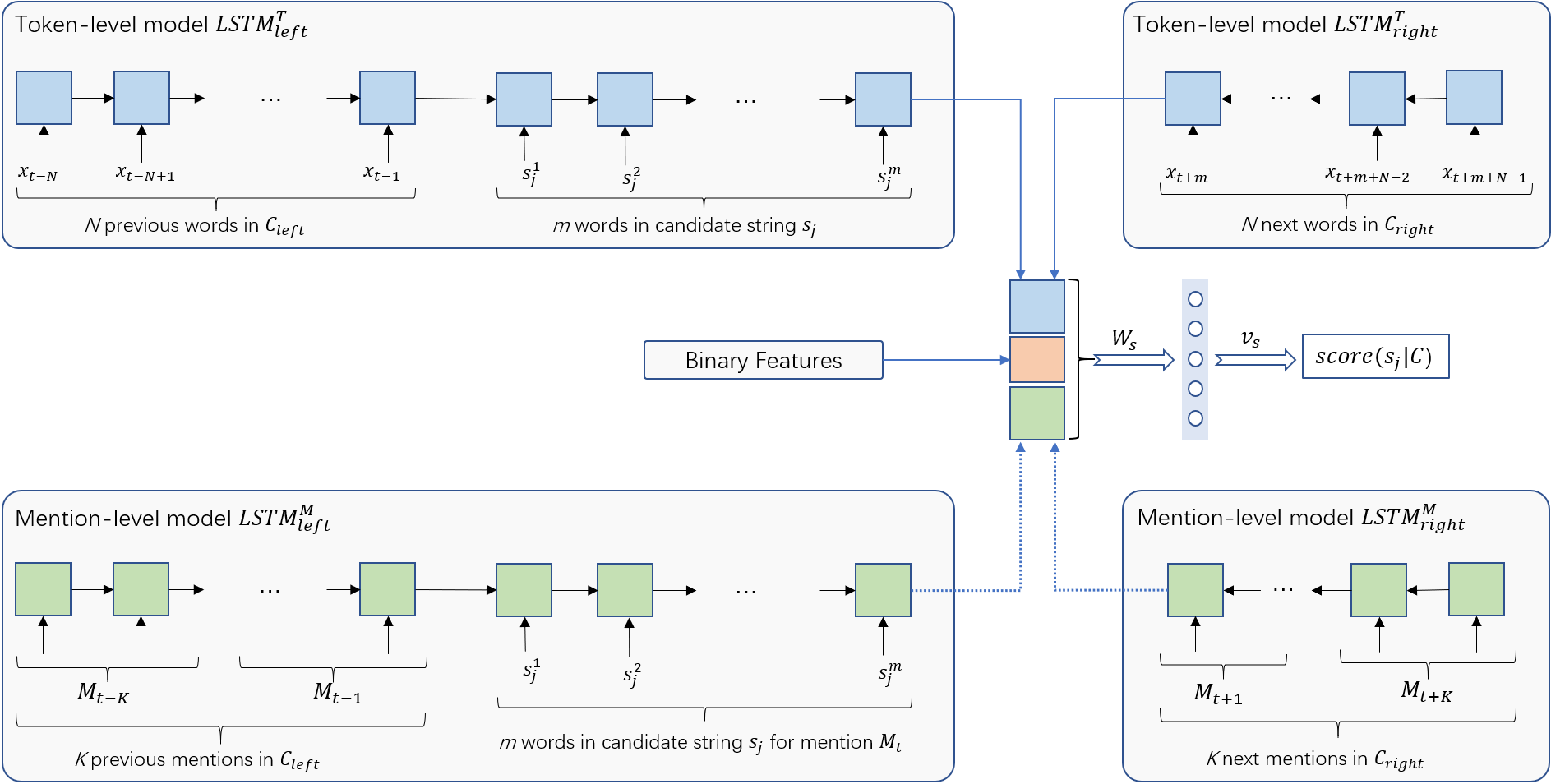}
	\caption{Computation of $score(s_j)$ using token-level and mention-level LSTMs for left and right contexts.} \label{fig:architecture}
\end{figure*}

Figure~\ref{fig:architecture} shows the neural architecture trained to compute the ranking $score(s_j|C)$. Given a mention position $t$ in the text, we split the context into a left context $C_{left}$ and a right context $C_{right}$, i.e. $C = [C_{left}, C_{right}]$, which are processed sequentially using two LSTM networks \citep{kn:Hochreiter97}:
% The left context refers to all the word tokens and mentions appearing to the left of mention position $t$, whereas the right context refers to all the word tokens and mentions appearing after position $t$. 
% The left and right contexts are processed sequentially using two LSTM networks \cite{kn:Hochreiter97}:
\begin{itemize}[topsep=0cm,itemsep=0cm,parsep=0cm,leftmargin=0.4cm]
	\item[$\circ$] {\bf Token-level}: $LSTM^T_{left}$ processes the $N$ tokens in the left context $C_{left}$ from left to right, followed by the tokens in the candidate string $s_j$. $LSTM^T_{right}$ processes the $N$ tokens in the right context $C_{right}$ from right to left, stopping right after the string $s_j$. The final states from the two LSTMs are concatenated to produce a token-level representation $h^{(T)}$. \textcolor{black}{In the example above, if $N=5$ and $s_j$="Nick", then $LSTM^T_{left}$ will process $\langle$",", "to", "his", "job", ".", "Nick"$\rangle$ in this order, whereas $LSTM^T_{right}$ will process $\langle$"every", "city", "the", "to", "drives"$\rangle$ in this order.}
	\item[$\circ$] {\bf Mention-level}: $LSTM^M_{left}$ processes the tokens from the $K$ mentions in the left context $C_{left}$, from left to right, followed by the tokens in the candidate string $s_j$. $LSTM^M_{right}$ processes the tokens from the $K$ mentions in the right context $C_{right}$, from right to left. The final states from the two LSTMs are concatenated to produce a mention-level representation $h^{(M)}$. \textcolor{black}{In the example above, if $K=2$ and $s_j$="Nick", then $LSTM^M_{left}$ will process $\langle$"Phil", "<sep>", "his", "<sep>", "Nick"$\rangle$ in this order, whereas $LSTM^M_{right}$ will process $\langle$"<pad>", "<sep>", "Emily"$\rangle$ in this order.}
\end{itemize}
\textcolor{black}{The use of LSTMs to process the context words and mentions is motivated by the fact that sequential information is essential for mention selection. For example, it is important to know whether a pronoun was used most recently for a different entity in the current sentence vs. two sentences before, because a pronoun should have a higher likelihood to be selected if the later were true, as it is less likely to create ambiguity while maintaining naturalness. Theoretically, when augmented with the binary features described below and when using a large value for the context size $N$, the token-level LSTM should be able to capture the relevant mention-level information. However, in practice, LSTMs can still "forget" relevant information from across large spans of text. Running the second LSTM only over the mention sequence is intended to alleviate this. The ablation results shown in Table~\ref{table:ablation} later in Section~\ref{sec:evaluation} further validate this design choice.}
%, whereas the comparative results shown in Table~\ref{table:performance-different-evaluation} demonstrate the importance of using sequential information, as LSTM models substantially outperform all decision-tree models.}

\textcolor{black}{Because the mention-level LSTM processes tokens from mentions, and because mentions may have a variable number of tokens, a special token $\langle sep \rangle$ is added before each mention to mark mention boundaries.}
% in the mention-level LSTM to indicate mention boundaries, as well as before the current mention ${M_t}$ in the token-level LSTM.
Since at test time the models will be run from left to right, this means that the strings for the relevant mentions at the right of the current mention are unknown. As such, they are replaced with a special token $\langle unk \rangle$ that receives its own embedding. For each token that is processed by the token-level and mention-level LSTM, we use as input their contextualized embeddings provided by BERT \cite{kn:Jacob18}. Additionally, for the mention-level LSTMs, the BERT embedding for each token is augmented with a vector of binary features $\phi_t^b$ computed as follows:
\begin{enumerate}
    \item Whether the mention belongs to the focus entity.
    \item A one-hot vector for the distance $d$ in tokens between the current mention and the previous mention. A binary feature is created for each of the following intervals: $d \leq 5$, $5 < d \leq 10$, $10 < d \leq 15$, $15 < d \leq 20$, $20 < d \leq 25$, and $d > 25$.
\end{enumerate}
These two features aim to help the system learn to avoid ambiguity and repetition. For example, if the mention $M_{t-1}$ right before ${M_t}$ is expressed as a pronoun, it does not refer to the focus entity, and the distance between the two mentions is small, then the system should learn to use a noun phrase for $M_t$ in order to avoid referential ambiguity.

In addition to the states computed by the LSTMs, we also use a vector $\phi^b$ of binary features, as follows:
\begin{enumerate}%[topsep=0cm,itemsep=0cm,parsep=0cm]%,leftmargin=0.4cm]
	\item Whether ${M_t}$ is the first or second mention in the coreference chain for $E$. 
	\item A one-hot vector for the mention length, i.e. the number of words contained in $s_j$. A separate binary feature is created for each length in $[1,5]$ and one binary feature for lengths greater than 5.
	\item Whether the candidate string $\mathit{s_j}$ for mention $M_t$ has already been used for a previous mention in the same coreference chain.
	\item Whether $s_j$ has been used for the previous mention $M_{t-1}$ in the coreference chain for entity $E$.
	\item Whether $M_t$ is the subject or object of a verb.
\end{enumerate}
The first two mentions in a coreference chain are usually longer noun phrases. Features 1 and 2 are expected to help the system learn to choose longer noun phrases for the first two mentions. Through features 3 and 4, the system is expected to learn to avoid repetition by using different strings. \textcolor{black}{The information of whether the mention is a subject or object provided through feature 5 should help the system to choose grammatically correct mention strings, for instance avoid selecting the object pronoun "him" if the mention is in a subject position, as in "{\it Bill} told Jake to go to work". Note that in this example the original string {\it Bill} on its own cannot give any information as to whether the original mention string was in a subject or object position. Note that with the exception of feature 2, all binary features rely on coreference or syntactic dependency information, which is unlikely to be learned implicitly by the LSTM on such a small dataset, hence the utility of encoding these features explicitly.}

The states from the token-level and mention-level LSTMs are concatenated with the binary features to create an overall feature vector $\phi(s_j, C) = [h^{(T)} | h^{(M)} | \phi^b]$. \textcolor{black}{Following the ranking approach of \cite{JMLR:v12:collobert11a}, this feature vector is then fed into a fully connected network with one hidden layer $(W_s, b_s)$ using {\it tanh} as non-linearity, followed by a linear output transformation $(v_s)$ into a final ranking score as follows:}
\begin{equation}
score(s_j | C) = v_s^T \tanh(W_s\phi(s_j, C) + b_s) %\nonumber
\end{equation}

\section{Datasets for PoV Change in English}
\label{sec:datasets}

Training deep learning models, such as the neural mention selection architecture that we propose for this task, requires a large number of training examples. However, manually annotating the corresponding large number of documents with PoV changes is prohibitive. Instead, we propose to use existing coreference resolution corpora to create training examples for the mention selection part of the PoV change task, as described in Section~\ref{sec:conll-data} below. Separately, we also create a benchmark dataset containing manual annotations of PoV changes, to be used only during testing, as described in Section~\ref{sec:pov-data}.

\subsection{Using Coreference Chains for Training}
\label{sec:conll-data}

The CoNLL-2012 corpus \citep{kn:pradhan-etal-conll-st-2012-ontonotes} is a widely used dataset for the training and evaluation of machine learning approaches to coreference resolution. \textcolor{black}{The dataset adds a coreference annotation layer to the OntoNotes corpus \cite{hovy-etal-2006-ontonotes,weischedel2011ontonotes}. The coreference layer annotates general anaphoric relations covering a rich set of entities and events that are not limited to noun phrases or a limited set of entity types. In this paper, we use the English portion of the dataset, which is comprised of documents from various genres of text, ranging from news, to conversational telephone speech, weblogs, Usenet newsgroups, broadcast, and talk shows.} For each document in the corpus, we select only the coreference chains that correspond to person entities and have the $3^{rd}$ person PoV. Each of these coreference chains can be used as the focus entity, \textcolor{black}{which is assumed to have been transformed from $1^{st}$ to $3^{rd}$ person PoV}, while the remaining chains that agree in number and gender are used as confounding entities. \textcolor{black}{Since input documents for the PoV change task should contain at most 1 entity with $1^{st}$ or $2^{nd}$ person PoV, as per the task definition ins Section~\ref{sec:definition}, this requires removing from the CoNLL corpus all documents that contain deictic PoVs}. \textcolor{black}{We use the official partition of the CoNLL-2012 corpus into training, validation, and testing. Upon filtering out documents with deictic PoVs, the statistics of the remaining documents in terms of person entities and their mentions are shown in the top 3 rows of Table~\ref{tab:statistics-conll}. Overall, about half of the original CoNLL dataset is used in our experiments.}
Given a coreference chain $E = \{m_1, m_2, ..., m_T\}$, we create $S(E)$ by collecting the unique strings that are used across all the mentions in the chain. Then the mention selection part of the PoV task can be reduced to selecting, for each position $t$, the actual string in $S(E)$ that was used to express mention $m_t$.
% This is done by training a ranker over the strings in $S(E)$ such the string that was actually used for mention $M_t$ is given a higher ranking score than the other strings in $S(E)$, as described in Section~\ref{sec:neural}.
%More details about the implementation of the ranker and the training procedure will be given in Section~\ref{sec:system}.
\begin{table}
\small
	\centering		
	\begin{tabular}{| l | r | r | r | r | r |}
		\hline
		Dataset & Entities & Mentions & Men / Ent
		& Docs & Words \\
		\hline
		\hline
		CoNLL-train & 1,952 & 11,727 & 6.0 & 1,143 & 603,885 \\
		CoNLL-dev & 267 &  1,446 & 5.4 & 154 & 80,181 \\
		CoNLL-test & 253 &  1,496 & 5.9 & 124 & 73,582 \\
		%\hline
		PoV-change & 300 & 8,682 & 28.9 & 21 & 84,312 \\
		% PoV-dev & 78 &  1,979 & 7 & 22,005 \\
		% PoV-test & 196 &  5,783 & 14 & 66,942 \\
		\hline
	\end{tabular}
%	\textcolor{black}{\caption{The number of person entities and mentions, average mentions per entity (Men / Ent). as well as the number of documents and words in each dataset.}	}
	\caption{The number of person entities and mentions, average mentions per entity (Men / Ent). as well as the number of documents and words in each dataset.}
	\label{tab:statistics-conll}
\end{table}

\subsection{The PoV Change Dataset}
\label{sec:pov-data}

\textcolor{black}{To enable the evaluation of various approaches to PoV change task, we created a separate corpus of documents in which the PoV change was done manually for the main character in each document. The major criteria for selecting documents to include in this benchmark dataset were as follow:
\begin{enumerate}
    \item The text should contain a first- or second-person narrative with a sufficiently large number of mentions (at least 10), such that a mention selection component would have to make non-trivial decisions. This typically means generating names, nouns, and pronouns that refer to the corresponding entity unambiguously, while maintaining the naturalness of the text.
    \begin{itemize}
        \item Because dialogues an direct speech are left unchanged (Section~\ref{sec:interactions}), $1^{st}$ and $2^{nd}$ person pronouns that appear in such contexts are not counted.
    \end{itemize}
    \item To enable the evaluation of a system's capability to avoid referential ambiguity, the text should also contain one or more confounding entities, with a non-trivial total number of mentions (at least 5). This means that the narrative should have one or more characters of the same gender as the focus character, which for the first-person PoV would be the narrator.
    \item The corpus should contain texts written in different styles, in order to capture a diverse set of referential devices and situations. 
\end{enumerate}
Upon a thorough exploration of the texts in the original (full) CoNLL corpus, we found that only 3 texts satisfied these criteria: one from the training portion (a2e\_0015), one from development (c2e\_0040), and one from test (c2e\_0029). In the broadcast news part of the corpus, for example, first and second person pronouns appear solely in dialogues in direct speech. Telephone conversations, talk shows, and broadcast conversations consist solely of dialogues, whereas the only situations in which first and second person points of view appeared in news and magazine texts were in direct speech. The only 3 texts that satisfied the corpus inclusion criteria were all found in the Usenet newsgroups part of CoNLL. In Usenet too the $1^{st}$ and $2^{nd}$ person pronouns are predominantly within dialogues and direct speech, although the dialogues are often implicit: a newsgroup post either responds to the author of an earlier post or addresses all readers of the newsgroup using $2^{nd}$ person pronouns. These types of narratives also often contain numerous {\it imperative statements}, e.g. "Excuse me if I am unable to answer", "Do not misunderstand me" (c2e\_0024), "I wish that you were more logical, my dear brother" (a2e\_0011), "If you think i am exaggerating things to scare people, then shut up", "Think of whether you have actual power to" (c2e\_0034), "Please guard your moral bottom line"(c2e\_0033), or {\it rhetorical questions}, e.g. "What do you need?", "Do you get excited?", "Are you worried?", "Are you determined?" (c2e\_0012). For both imperative statements and rhetorical questions it is unclear whether a PoV change would make sense.}

\textcolor{black}{Since the training part  is used to train the mention selection methods, from the CoNLL corpus we manually annotated only the document c2e\_0040 from development and c2e\_0040 from test that satisfied the corpus inclusion criteria (making sure to eliminate c2e\_0040 from development during hyper-parameter tuning). We completed the PoV change corpus by manually annotating an additional set of 19 texts, which were taken from the sources listed in Table~\ref{tab:articleinfo1}.
}
In 18 texts the PoV was changed from $1^{st}$ to $3^{rd}$ person, whereas in the text from \cite{kn:Melanie} the PoV was changed from $2^{nd}$ to $3^{rd}$ person. \textcolor{black}{The inclusion of only one text with a $2^{nd}$ person PoV is a consequence of the fact that in narratives the second person point of view is much less common, compared to first person.} The manual annotation was done by a graduate in Linguistics.

\begin{table}[t]
	\footnotesize	
	\centering
	\begin{tabular}{|r|c|rrr|rrr|}
		\hline%
		Source & Genre / Type & Cf & M$_{cf}$ & MC$_e$ & P & R & F$_1$ \\
		\hline
		\hline
		\cite{kn:Chu13} & Fiction / Short story & 14 & 33.4 & 3.5 & 70.3 & 73.8 & 72.0\\
		\cite{kn:Ted91} & Fiction / Short story & 7 & 33 & 4.9 & 59.5 & 65.4 & 62.3\\
		\cite{kn:Michael77} & Non-fiction / Book excerpt & 37 & 10.5 & 3.4 & 71.7 & 72.0 & 71.8\\
		\cite{kn:Flynn05} & Fiction / Book excerpt & 34 & 13.5 & 3.3 & 73.8 & 76.5 & 75.1\\
		\cite{kn:Bryson10} & Non-fiction / Book excerpt & 13 & 17.7 & 2.6 & 71.4 & 72.2 & 71.8\\
	    \cite{kn:Pollan18}  & Fiction / Book excerpt & 17 & 12 & 2.8 & 83.9 & 81.7 & 82.8\\
		\cite{kn:Solnit13} & Non-fiction / Book excerpt & 20 & 20.2 & 3 & 73.8 & 72.7 & 73.2\\
		\cite{kn:Sofia} & Fiction / Short story & 15 & 19.9 & 3.8 & 74.0 & 74.5 & 74.2\\
		\cite{kn:Vonnegut99} & Fiction / Book excerpt & 18 & 11.1 & 3.7 & 70.3 & 67.0 & 68.6\\
		\cite{kn:Issac} & Fiction / Book excerpt & 7 & 41.3 & 4.1 & 73.5 & 72.5 & 73.0\\
		\cite{kn:Scaachi} & Non-fiction / Short story & 9 & 16.2 & 3.1 & 73.5 & 74.4 & 73.9\\
		\cite{kn:Melanie}$(*)$ & Non-fiction / Web article & 4 & 57.5 & 5.0 & 62.8 & 69.2 & 65.8\\
		CoNLL-c2e-0029 & Non-fiction / Web forum & 6 & 9.7 & 3.6 & 74.2 & 72.0 & 73.1 \\
		CoNLL-c2e-0040 & Non-fiction / Web forum & 1 & 4.0 & 2.5 & 82.8 & 80.0 & 81.4 \\
		\hdashline[1pt/1pt]
		\cite{kn:Shoshana} & Non-fiction / Web article & 9 & 17.7 & 4.2 & 71.5 & 73.6 & 72.5\\
		\cite{kn:Toni} & Fiction / Short story & 16 & 11.4 & 3.4 & 74.3 & 77.7 & 76.0\\
		\cite{kn:Biss09} & Non-fiction / Book excerpt & 10 & 9.1 & 3.1 & 68.1 & 67.0 & 67.5\\
		\cite{kn:Mary} & Fiction / Short story & 9 & 6.3 & 2.9 & 81.9 & 80.5 & 81.2\\
		\cite{kn:Atwood98} & Fiction / Book excerpt & 17 & 27.4 & 3.6 & 68.9 & 70.2 & 69.5\\
		\cite{kn:Harrison} & Non-fiction / Short story & 9 & 21.4 & 3.8 & 74.0 & 73.2 & 73.6\\
		\cite{kn:Sartre69} & Fiction / Book excerpt & 7 & 16.7 & 4.1 & 70.0 & 68.3 & 69.1\\
		\hline
	\end{tabular}
	\caption{The sources used for the PoV change dataset. Cf = total number of confounding entities; M$_{cf}$ =  average number of mentions per confounding entity; MC$_e$ = average number of mention candidates per entity; \textcolor{black}{P, R, and F$_1$ are the precision, recall, and F-measure of the end-to-end PoV change system on each document.}}
	\label{tab:articleinfo1}
\end{table}

The texts were selected to provide a mix of fiction and non-fiction, as well as a combination of short stories, web articles, and book excerpts, and not be dominated by dialogue. For instance, {\it Sweetness} \cite{kn:Toni} is an emotional confession of a mulatto mother about how she treated her black daughter. She often addresses the reader directly, which makes it hard to change the PoV without modifying more of the text.
%For example:
%\begin{enumerate}[topsep=0cm, leftmargin=0.4cm]
%	\item[$\circ$] \emph{Some of you probably think it’s a bad thing to group \colorbox{light-gray}{ourselves} $\rightarrow$ \colorbox{light-gray}{themselves} according to skin color—the lighter the better—in social clubs, neighborhoods, churches, sororities, even colored schools.}
%\end{enumerate}
{\it Mother Night} \cite{kn:Vonnegut99} is a novel written as a fictional memoir of a former Nazi propagandist named Howard Campbell. The narrative alternates between present and past and the narrator’s voice is heavily present, which makes this a more complex text. For example:
\begin{enumerate}[topsep=0cm,itemsep=0pt,leftmargin=0.4cm]
	\setcounter{enumi}{1}
    \item[$\circ$] \emph{When \colorbox{light-gray}{he} $\rightarrow$ \colorbox{light-gray}{Kraft} opened up, \colorbox{light-gray}{I} $\rightarrow$ \colorbox{light-gray}{Campbell} saw that he was a painter. There was an easel in the middle of his living room ...} %with a fresh canvas on it, and there were stunning paintings by him on every wall.}
    \item[$\circ$] \emph{When I talk about Kraft, alias Potapov, I’m a lot more comfortable than when I talk about Wirtanen, who has left no more of a trail than an inchworm crossing a billiard table. Evidences of Kraft are everywhere.}
    % At this very moment, I’m told, Kraft’s paintings are bringing as much as ten thousand dollars a piece in New York.}
\end{enumerate}
In the first paragraph above, the writer recounts Campbell’s story, using the past tense. Then, in the second paragraph, he switches to the present tense, providing additional information about one of the characters and expressing his feelings regarding him. As this is clearly the voice of the author, it needs to stay in the $1^{st}$ person.

To enable assessing the performance of mention selection, these articles were also manually annotated with coreference chains and verb conjugation changes. The PoV change dataset is used solely for evaluation purpose and was originally partitioned into validation and testing, as shown in Table \ref{tab:articleinfo1} in the top 14 rows and bottom 7 rows, respectively. \textcolor{black}{Eventually, as tuning on the development portion of the PoV dataset did not lead to better mention selection accuracy when compared with tuning on CoNLL development data, the PoV change results were computed over all 21 documents. The statistics of the full PoV dataset are shown in the last line of Table~\ref{tab:statistics-conll}. While the number of texts may seem small, the number of words is comparable (slightly larger) than in CoNLL-test. This is because the PoV documents are much larger than those in CoNLL. On average, a PoV document has 4,015 words, whereas the CoNLL-test document is on average only 593 words long. Out of the 19 PoV documents that are not from CoNLL, the smallest document is \citep{kn:Biss09} with 1,106 words, whereas the largest is \citep{kn:Chu13} with 7,611 words. Furthermore, the number of person entities is larger in PoV than in CoNLL-test, whereas the total number of mentions in PoV is more than 3 times the number of mentions in CoNLL-test. Coupled with the fact that PoV documents are much longer, this results in an average number of mentions per person entity in PoV that is almost 5 times larger than in CoNLL-test. The longer coreference chains in the PoV dataset are important in that they enable evaluating the ability of PoV change systems to select mention strings that are not repetitive or ambiguous over larger spans of texts, and in the presence of multiple confounding entities.}

Given an input text and a focus entity, there may be multiple solutions to the task of PoV change, each a with different trade-off between the manageable ambiguity and naturalness criteria. To get a sense of the potential differences, a fragment from the autobiography of \cite{kn:Kurosawa83} was annotated independently by two annotators. There are 169 focus entity mentions in the annotated fragment. In 121 mention conversions both annotators used the same pronouns, whereas in 22 mention conversions both used the same noun phrases -- including one situation where both used a noun plus pronoun combination, but one chose to use anaphora and the other cataphora. There are 26 cases where one chose to use pronouns and the other chose to use noun phrases, as in the example below:

\hangindent=0.4cm \emph{\{I\} $\rightarrow$ \{Akira\} had been one year old at the time. The dimly lit place where \{I\} $\rightarrow$ \{{\bf he}\} / \{{\bf the baby}\} sat in a tub lodged between two boards was the room that served as both kitchen and bath in the house where \{my\} $\rightarrow$ \{his\} father was born.}

\noindent In most cases the two annotators chose to use noun phrases during transition between past and present, then for clarity, and lastly for stylistic reasons. Overall, the inter-annotator agreement was 84.6\%.

\section{Experimental Evaluation}
\label{sec:evaluation}

The training portion of the CoNLL-2012 corpus is used to train two LSTM-based models: a model that uses only the token-level LSTMs, and a model that uses both the token-level and mention-level LSTMs. \textcolor{black}{The size of the hidden state of LSTM cell is set to 50, whereas the size of the hidden layer of the fully connected network that computes the ranking score is set to 100.} We use $N = 50$ words and $K = 10$ mentions for both the left and right contexts. If there are less than $N$ words or $K$ mentions available, we use a special token $\langle pad \rangle$ with its own embedding to pad the sequence.
% \textcolor{black}{A special token $\langle sep \rangle$ is used before each mention in the mention-level LSTM to indicate mention boundaries, as well as before the current mention ${M_t}$ in the token-level LSTM.}

The models are implemented in PyTorch. The learning rate, the margin $\gamma$, and the dropout rates are tuned using grid search on the validation portion of CoNLL-2012, which is also used for early stopping. The models are then tested separately on the corresponding test portion of CoNLL and the PoV dataset. \textcolor{black}{Because models that were originally tuned on the PoV development portion obtained the same results as models tuned on CoNLL-dev, we combined the development portion of PoV with the test portion and report results on the entire set of 21 documents using models trained and tuned solely on CoNLL.}
%We use an early-stopping strategy on the validation portion of the corresponding dataset.
Optimization is performed with the Adam algorithm \cite{kn:Kingma15}. Testing is done in auto-regressive mode, by doing mention selection from left to right. \textcolor{black}{The models were tuned and trained on a desktop computer with an Intel Core i7-8700K CPU @3.70GHz with 16GB memory, and an NVIDIA GeForce GTX 1080 with 8GB memory. Training the tuned model on this platform took 16 hours and 50 minutes.} 
% Therefore, we need to obtain the BERT embedding for each token every time running the program on the testing data.

\subsection{Automatic Evaluation}

Models are evaluated using precision (P), recall (R), and F$_1$ score with respect to both the mention and verb conversions, using the manual annotations as ground truth. \textcolor{black}{Thus, if the system changed a total of $N$ words, and $N_1$ of them were correctly changed, then precision is calculated as $N_1 / N$. If the ground truth specifies that $N_2$ words should have been changed, then recall is calculated as $N_1 / N_2$. The F$_1$ measure is then calculated as the harmonic mean of precision and recall. For the mention selection evaluations, since the number of mentions to change is known, i.e. $N$ is always the same as $N_2$, we report accuracy as the percentage of mentions that were changed to the ground truth string.}
%The performance are reported for the testing portions of the CoNLL and PoV dataset. 
Table \ref{table:performance-different-evaluation} reports two types of test results: {\it End-to-End} refers to the performance of the end-to-end system, where mention detection, coreference resolution, verb conjugation, and generation of $S(E)$ are done automatically by the system; {\it Mention Selection} refers to the performance of the mention selection component, using the manual annotations for everything else. \textcolor{black}{Since the CoNLL dataset does not have the verb conjugation annotation for the $1^{st}$ person PoV, we report only the mention selection performance for it. Furthermore, the End-to-End performance of the full {\it LSTM: Token + Mention} model on each document in the PoV dataset is presented in the last 3 columns of Table~\ref{tab:articleinfo1}.}

\begin{table}[t]
	\small
	\centering
	\begin{tabular}{|l|c|c|ccc|}
		\cline{2-6}
	  	\mc{} & \multicolumn{2}{c|}{Mention Selection} & \multicolumn{3}{c|}{PoV End-to-End} \\
		 \hline
	    Model & CoNLL & PoV & P & R & F1\\
		\hline
		LSTM: Token + Mention & {\bf 74.1}  & {\bf 75.2} & {\bf 71.6} & {\bf 73.0} & {\bf 72.3}\\
		LSTM: Token & 70.9  & 73.4 & 69.7 & 71.1 & 70.4\\
		\hline%
		 Decision Trees & 52.7 & 57.2 & 49.1 & 50.0 & 49.5\\
		Random Forests & 57.7 & 59.8 & 54.7 & 55.8 & 55.2\\
		 Gradient Boosted Trees & 64.8 & 62.9 & 57.2 & 58.3 & 57.7\\
		 \hline%
		 Random Selection & 33.9 & 33.6 & 21.6 & 22.1 & 21.8\\
		Only Pronouns & 28.6 & 42.5 & 37.1 & 37.8 & 37.4\\
		Most Common String$^*$ & 45.4 & 48.8 & 37.1 & 37.9 & 37.5\\
        \hline
	\end{tabular}
	\caption{Comparative model performance (\%) in terms of accuracy for Mention Selection, Precision (P), Recall (R), and F1 score for End-to-End.}
	\label{table:performance-different-evaluation}
\end{table}

We also present the system performance when mention selection is done using 2 simple baselines and 3 trained decision tree-based approaches. The {\it Random Selection} baseline chooses a string from $S(E)$ at random. The {\it Only Pronouns} baseline uses only pronouns, making sure that they satisfy gender, number, and case agreement. We also present the results of a {\it Most Common String} approach which uses only the most common string across all mentions of $E$. Note that this approach cannot be used in practice because it needs to know the true counts of strings from $S(E)$, which is unknown at test time. We also implemented mention selection using 3 decision tree-based approaches: {\it Decision Trees, Random Forests}, and {\it Gradient Boosting Trees}. A set of binary features are calculated for each of the 10 mentions before and after the current mention, as follows:
\begin{enumerate}[topsep=0cm,itemsep=0cm,parsep=0cm]
\item The distance $d$ in words between the mention and the current mention. A separate binary feature is created for each of the following distance intervals: $d \in [0, 5]$, $d \in (5, 10]$, $d \in (10, 15]$, $d \in (15, 20]$, $d \in (20, 25]$, $d > 25$.
\item If the mention string is a pronoun, proper noun or common noun.
\item If the mention belongs to the same entity as the current mention.
\item If the mention belongs to the same sentence as the current mention.
\item If the mention has the same number and gender as the current mention.
\item If the mention string is the same as the string from $S(E)$ used for the current mention.
\item The mention length $l$ in words. A separate binary feature is created for each value of $l \in [1,5]$ and for $l > 5$.
\item If the mention is $\langle pad \rangle$.
\end{enumerate}
A similar set of binary features are extracted for the 10 mentions that appear before the current mention and belong to the same entity as the current mention. A BERT score is calculated for each mention string from $S(E)$, and used as a feature to indicate how likely it is for that string to appear in the context for the current mention. Other features used for the current mention include its part of speech, its length, and whether it is the subject or object of a verb. Finally, we use the original mention string to narrow down the choices in $S(E)$, as described in Section~\ref{sec:neural}. Note that the binary features used by the decision tree-based models subsume all the binary features used in the mention-level and token-level LSTMs.

The decision tree-based models are trained as binary classifiers. For each $s \in S(E)$, if $s$ is the correct string for the current mention, then the label is positive, otherwise negative. During testing, for a given mention position, the string $s$ from $S(E)$ that obtains the highest probability to be positive is selected as the correct mention string. \textcolor{black}{We used decision trees as a baseline as we were also interested in the performance of ML models that can ultimately be encoded as rules. Furthermore, ensembles of decision trees such as Random Forests and Gradient Boosted Trees are known to be very effective ML models.}

The results in Table~\ref{table:performance-different-evaluation} demonstrate the importance of modeling the entities separately through the mention-level LSTMs, as the full Token + Mention model outperforms the Token model on both datasets. The mention selection performance on the PoV change dataset is better than the end-to-end performance, which is expected due to error propagation from previous stages in the pipeline. The proposed LSTM-based models significantly outperform the baselines, and also perform better than the decision tree approaches. The mention selection performance on the PoV dataset is slightly better than on CoNLL. This is because the CoNLL performance is calculated solely based on mention conversions, while PoV performance numbers are based on both mention selection and verb conjugations, which have a relatively higher accuracy. All trained models were tuned on the CoNLL validation data. We also tuned on PoV validation data, however that lead to very similar performance.
% And we can save all the PoV data for testing and only validate on the CoNLL data in future.

Table~\ref{tab:articleinfo1} also shows the precision (P), recall (R), and F1 measure of the End-to-End pipeline for each article. The key factor affecting the performance appears to be the average number of mention candidates per entity MC$_e$, with values greater than 3.8 leading to a more marked decrease in performance. \textcolor{black}{Figure~\ref{fig:acc-vs-se} shows the accuracy of the mention selection model {\it LSTM: Token + Mention} as a function of the size of the set of candidate entities $S(E)$. Overall, as expected, the mention selection accuracy trends down as the size of $S(E)$ increases.
\begin{figure}[ht]
%\vspace*{-20pt}
\centering
	\includegraphics[width=0.8\textwidth]{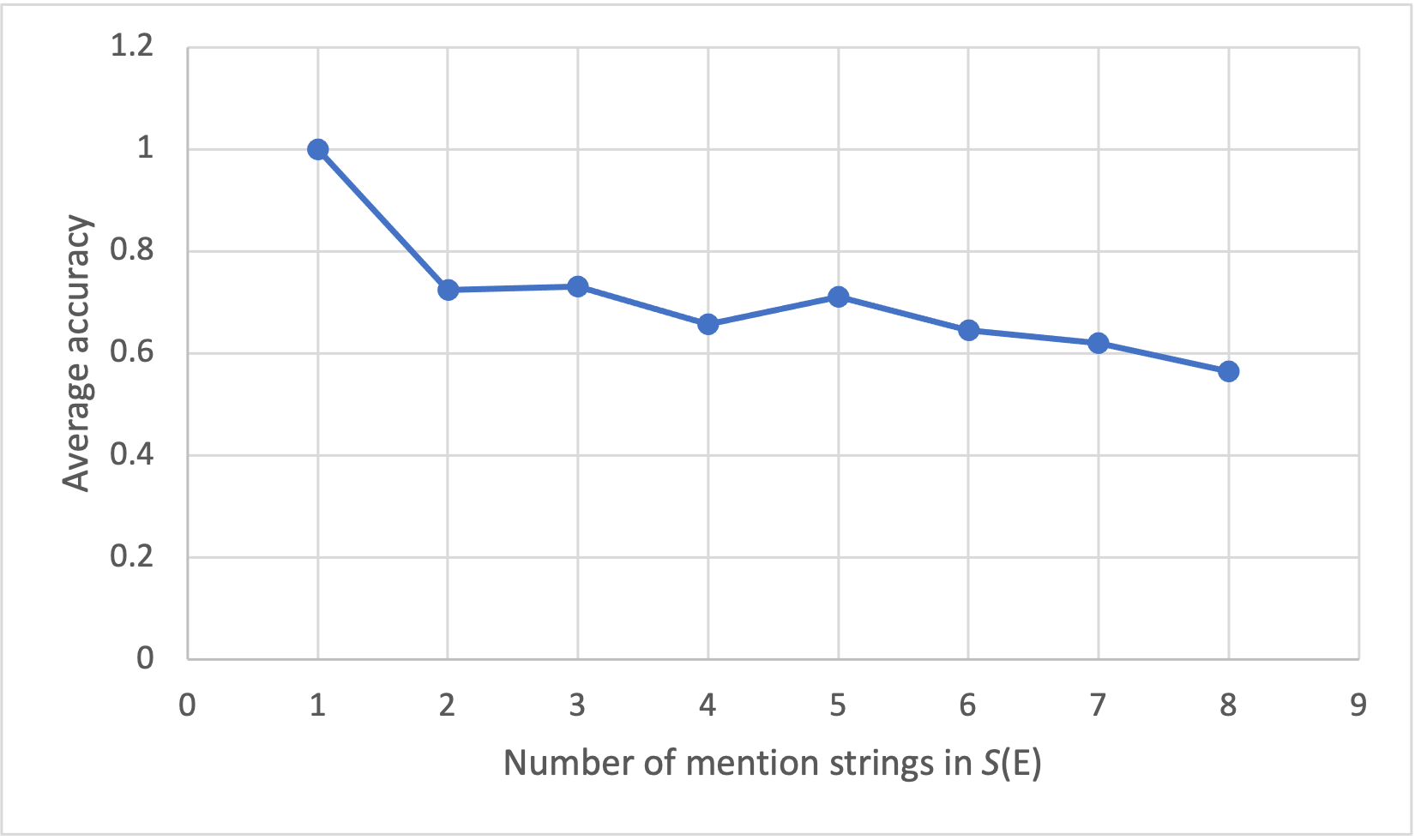}
	\caption{Accuracy of mention selection for different sizes of $S(E)$.} \label{fig:acc-vs-se}
\end{figure}
}

% The average number of mentions per confounding entity M$_{cf}$ also affects the performance: when two articles have the same average number of mention candidates per entity, then the article with a smaller average number of mentions per confounding entity $A(CE_M)$ usually has better performance.

Table~\ref{table:ablation} shows the results of 5 ablation tests, demonstrating that all components of the mention selection architecture are beneficial to the final End-to-End performance: the token-level LSTM, the mention-level LSTM, the token-level binary features $\phi_t^b$, and the overall vector of binary features $\phi^b$. \textcolor{black}{We ran statistical test of significance using the {\it one-tailed paired T-test} over the F$_1$ results for the 21 documents in the PoV change dataset, i.e. 21 results for each of the systems in the comparison. The following comparisons were highly statistically significant ({\it p-value} $< 0.01$):
\begin{itemize}
    \item {\it LSTM: Token + Mention} is better than {\it LSTM: Token}, i.e. the mention-level LSTM is useful.
    \item {\it LSTM: Token + $\phi_t^b$} is better than {\it LSTM: Token}, i.e. the token-level binary features are useful.
\end{itemize}
Furthermore, the following comparison was statistically significant ({\it p-value} $< 0.05$):
\begin{itemize}
    \item {\it LSTM: Token + Mention} is better than {\it LSTM: Mention}, i.e. the token-level LSTM is useful too.
\end{itemize}
}
% "Token with token-level binary features" represents the model with only token-level LSTMs, for which the BERT embedding of each token is augmented with three types of binary features: (1) Whether the token belongs to a mention of a person entity or not; (2) Whether the token belongs to the focus entity or not; and (3) A one-hot vector for the distance d in tokens between the current mention and the previous mention. A separate binary feature is created for each of the following distance intervals: $d \leq 5$, $5 < d \leq 10$, $10 < d \leq 15$, $15 < d \leq 20$, $20 < d \leq 25$, and $d > 25$.}

\begin{table}[t]
	\small
	\centering
	\begin{tabular}{|l|ccc|}
        \hline
	    Model & P & R & F$_1$ \\
		\hline
		LSTM: Token + Mention & {\bf 71.6} & {\bf 73.0} & {\bf 72.3}\\
		LSTM: Token + Mention $-$ $\phi^b$ & 71.3 & 72.8 & 72.0\\
		LSTM: Token + Mention $-$ $\phi_t^b$ & 71.3 & 72.7 & 72.0\\
		LSTM: Token & 69.7 & 71.1 & 70.4\\
		LSTM: Mention & 71.0 & 72.4 & 71.7\\
		LSTM: Token + $\phi_t^b$ & 71.1 & 72.5 & 71.8\\
		\hline%
	\end{tabular}
	\caption{Ablation results (\%) for the End-to-End scenario in terms of Precision (P), Recall (R), and F1 score.}
	\label{table:ablation}
\end{table}

Table \ref{table:error-source} shows the  performance of each component in the system pipeline. For each component, the evaluation is done under the assumption that the preceding components in the pipeline run without errors, e.g. for the verb identification results we assume that the mention detection and coreference resolution, on which verb detection identification depends, are entirely correct.
\begin{table}[t]
	\small
	\centering	
	\begin{tabular}{| l | c | c | c | }
		\hline
		Pipeline component & P & R & F1\\
		\hline
		Coreference resolution & 91.1 & 96.9 & 93.9\\
		Verb identification & 75.2 & 76.7 & 75.9\\
		Verb conjugation & 100.0 & 100.0 & 100.0\\
		Generation of $S(E)$ & 99.1 & 99.1 & 99.1\\
		\hline
	\end{tabular}
	\caption{Performance (\%) in terms of Precision (P), Recall (R), F1 score for each component in the pipeline.}
	\label{table:error-source}
\end{table}

Sample texts generated by the system are shown in Table~\ref{table:sample-texts}. Most of these example show that the system is able to make sophisticated mention selection decisions, where pronouns alternate with names and noun phrases in order to alleviate potential ambiguity introduced by confounding entities, while also preserving the naturalness of the text. In sentences 1 and 3, for instance, the system learns to not use the same third person pronoun for different entities mentioned in subject position in consecutive clauses.

\begin{table}[!t]
	%\small	
	\footnotesize
	\begin{tabularx}{\textwidth}{X}
%	\hline%
	\underline{{\it Focus Entity}: {\bf I (Campbell)}, {\it Confounding Entity}: {\bf George Kraft}} \ \newline
	1. When {\it $[$he$]_1$} $\rightarrow$ {\bf $[$Kraft$]_1$} opened up, {\it $[$I$]_2$} $\rightarrow$ {\bf $[$Campbell$]_2$} saw that {\it $[$he$]_1$} was a painter. There was an easel in the middle of {\it $[$his$]_1$} living room with a fresh canvas on it, and there were stunning paintings by {\it $[$him$]_1$} on every wall.\\
%	\hline%
    2. George Kraft was only one of {\it $[$his$]_1$} names. The real name of \colorbox{light-gray}{{\it $[$this old man$]_1$}} $\rightarrow$ \colorbox{pink}{{\bf $[$his$]_1$}} was Colonel Iona Potapov.\\
%		\hline%
%		\textcolor{black}{3. Well - when $[$Kraft$]_1$ opened $[$his$]_1$ door for \colorbox{light-gray}{$[$him$]_2$} $\rightarrow$ \colorbox{pink}{$[$Campbell$]_2$}, \colorbox{light-gray}{$[$Campbell$]_2$} $\rightarrow$ \colorbox{pink}{$[$he$]_2$} knew $[$his$]_1$ paintings were good.}\\
%	\hline%
	\hline
	%\hline
	\underline{{\it Focus Entity}: {\bf I (Nick)}, {\it Confounding Entity}: {\bf Nick's father}} \ \newline 	3. Even before {\it $[$he$]_2$} became homeless {\it $[$I$]_1$} $\rightarrow$ {\bf $[$Nick$]_1$} had heard whispers, sensed {\it $[$he$]_2$} $\rightarrow$ {\bf $[$his father$]_2$} was circling close, that they were circling each other; like planets unmoored.\\
%	\hline
	4. {\it $[$I$]_1$} $\rightarrow$ {\bf $[$Nick$]_1$} even knew what kind, a Town Taxi, a black and white. In {\it $[$my$]_1$} $\rightarrow$ {\bf $[$his$]_1$} early twenties, after {\it $[$I$]_1$} $\rightarrow$ {\bf $[$he$]_1$} dropped out of college and moved to Boston, {\it $[$I$]_1$} $\rightarrow$ {\bf $[$he$]_1$} would involuntarily check the driver of each that passed, uncertain what it would mean, what {\it $[$I$]_1$} $\rightarrow$ {\bf $[$he$]_1$} would do, if it was {\it $[$my$]_1$} $\rightarrow$ {\bf $[$his$]_1$} father behind the wheel.\\
%	\hline
	5. {\it $[$I$]_1$} $\rightarrow$ {\bf $[$Nick$]_1$} could have given {\it $[$him$]_2$} a key, offered a piece of {\it $[$my$]_1$} $\rightarrow$ {\bf $[$his$]_1$} floor. A futon. A bed. But {\it $[$I$]_1$} $\rightarrow$ {\bf $[$he$]_1$} never did. If {\it $[$I$]_1$} $\rightarrow$ {\bf $[$he$]_1$} let {\it $[$him$]_2$} $\rightarrow$ {\bf $[$his father$]_2$} inside {\it $[$I$]_1$} $\rightarrow$ {\bf $[$he$]_1$} would become {\it $[$him$]_2$}.\\
%	\hline
    6. If you asked {\it $[$me$]_1$} $\rightarrow$ {\bf $[$him$]_1$} about {\it $[$my$]_1$} $\rightarrow$ {\bf $[$his$]_1$} father then - the years \colorbox{light-gray}{{\it $[$his father$]_2$}} $\rightarrow$ \colorbox{pink}{\bf {$[$he$]_1$}} lived in a doorway, in a shelter, in an ATM - {\it $[$I$]_1$} $\rightarrow$ {\bf $[$Nick$]_1$} would say, ``Dead," {\it $[$I$]_1$} $\rightarrow$ {\bf $[$he$]_1$}'d say, ``Missing," {\it $[$I$]_1$} $\rightarrow$ {\bf $[$he$]_1$}'d say, "I don't know where he is."\\	
%	\hline
	\hline
	%\hline
	\underline{{\it Focus Entity}: {\bf I (Michael)}, {\it Confounding Entity}: {\bf Fritz (the guide)}} \ \newline
	7. At least on paper, nothing about {\it $[$the first guide$]_1$} {\it $[$I$]_2$} $\rightarrow$ {\bf $[$Michael$]_2$} chose to work with sounds auspicious.\\
%	\hline
	8. {\it $[$I$]_2$} $\rightarrow$ {\bf $[$Michael$]_2$} surprised {\it $[$myself$]_2$} $\rightarrow$ {\bf $[$himself$]_2$} by liking {\it $[$Fritz$]_1$} as much as {\it $[$I$]_2$} $\rightarrow$ {\bf $[$he$]_2$} did.\\
%	9. $[$Michael$]_2$ could just forget about the whole idea of being anywhere in range of a hospital emergency room. Then there was the fact that while $[$he$]_1$ was a Jew from a family that had once been reluctant to buy a German car, \colorbox{light-gray}{$[$this fellow$]_1$} $\rightarrow$ \colorbox{pink}{$[$Fritz$]_1$} was the son of a Nazi—a German in $[$his$]_1$ midsixties whose father had served in the SS during World War II.\\
	9. {\it $[$I$]_2$} $\rightarrow$ {\bf $[$Michael$]_2$} could just forget about the whole idea of being anywhere in range of a hospital emergency room. Then there was the fact that while {\it $[$I$]_2$} $\rightarrow$ {\bf $[$he$]_2$} was from a family that had once been reluctant to buy a German car, \colorbox{light-gray}{{\it $[$this fellow$]_1$}} $\rightarrow$ \colorbox{pink}{{\bf $[$Fritz$]_1$}} was a German in {\it $[$his$]_1$} midsixties whose father had served in the army during World War II.\\
%	\hline
	10. {\it $[$Fritz$]_1$} told \colorbox{light-gray}{{\it $[$me$]_2$}} $\rightarrow$ \colorbox{pink}{{\bf $[$Michael$]_2$}} what to expect if {\it $[$I$]_2$} $\rightarrow$ {\bf $[$he$]_2$} \dashuline{{\it were} $\rightarrow$ {\bf was}} to work with {\it $[$him$]_1$}.\\
	\hline
	\end{tabularx}	
	\caption{Sample texts generated by the system. Entity mentions are shown between brackets, indexed by their coreference chain. Highlighted in red are the system outputs that are different from the manual annotations shown in light gray.}
	\label{table:sample-texts}
\end{table}

There are also instances where, even though the system selection is different from the manual annotation, the result still sounds acceptable. In sentence 2, the system preferred ``his" over the annotated string ``this old man". Upon searching the training data, we found similar possessive constructions, as in ``And so this explanation of his is untruthful.", which may explain this mention selection. In sentence 6, even though the system preferred ``he" over ``his father" for the third mention, it is still an acceptable selection: while it increases the ambiguity, it makes the text sound more natural. 
For the confounding entity in sentences 7 and 9, the system selected ``the first guide" as in the manual annotation, but not ``this fellow", preferring to use the name instead. This is probably due to the fact that most of the common nouns used in the CoNLL training data are titles or occupations, such as "the president" or "the correspondent". However, even though the annotated "this fellow" is more coherent with the article's style, the system generated mention does decrease the ambiguity.

\subsection{Human Evaluation}
\label{sec:human-evaluation}

Evaluating the output of a PoV change system is complicated by the fact that there may be multiple solutions for choosing mention strings that achieve felicitous, non-ambiguous reference while also maintaining the naturalness of a sentence. Inspired by work in style transfer \citep{kn:Shen17,kn:rao-tetreault-2018-dear,kn:prabhumoye-etal-2018-style}, we conducted human-based evaluations to assess the quality of the {\it Manual}-ly annotated text vs. the system-generated text in the {\it Mention Selection} and {\it End-to-End} scenarios.
% The two major challenges for the task are generating {\it unambiguous} and {\it natural} text, as such they are the focus of the human evaluation. 
To obtain high quality annotations from Amazon Mechanical Turk (MTurk) workers, we followed \citet{kn:clark-etal-2018-neural} and only used US-located workers with a HIT approval rate greater than 95\% and a number of approved HITs greater than 1,000. Each worker was presented with a paragraph from a story in the PoV dataset, one sentence at a time. For each mention in the sentence, the worker was asked to map it to the corresponding discourse entity, i.e. its referent, by selecting from a predefined list of entities known to appear in the original version of that paragraph. To understand the context of the current sentence, the previous sentences were also shown to the worker. Each worker was asked to annotate three versions of the paragraph: Manual, System: Mention Selection, and System: End-to-End. The three versions were presented in random order.

To estimate referential ambiguity, each worker $w$ was asked to indicate how easy it was to map the mention string $m$ to its referent entity, using a rating $amb(w,m)$ ranging from 0: {\it Hard} (high ambiguity), to 1: {\it Moderate} (some ambiguity), to 2: {\it Easy} (no ambiguity). The workers also had the option of indicating whether the mention could not be mapped to any of the listed entities, using an {\it Impossible} rating. This can happen when a wrong string is used for that mention, mainly due to mistakes in coreference resolution that propagate in the construction of $S(E)$. \textcolor{black}{Below are examples illustrating ratings of entity disambiguation difficulty, which were provided as part of the instructions to the MTurk workers:\\
\begin{small}
{\bf Previous sentences}: The door to Reynolds' apartment is also open. Leon walks down the entryway to the living room, hearing a hyperaccelerated polyphony from a digital synthesizer.
\begin{enumerate}
    \item {\bf Current sentence}: Evidently, it's {\it Reynold's} own work; the sounds are modulated in ways undetectable to normal hearing, and even \underline{Leon} can't discern any pattern to them. \\
    \APLcirc{\Circle} Easy \Circle Moderate \Circle Hard \Circle Impossible\\
    {\bf Analysis}: The mention string "Leon" clearly refers to Leon. Correspondingly, there is no referential ambiguity.
    \item {\bf Current sentence}: Evidently, it's {\it Reynold's} own work; the sounds are modulated in ways undetectable to normal hearing, and even \underline{he} can't discern any pattern to them. \\
    \Circle Easy \APLcirc{\Circle} Moderate \Circle Hard \Circle Impossible\\
    {\bf Analysis}: The pronoun "he" may refer to either Leon or Reynold. However, according to the context, it is Reynold's own work, therefore it should be Leon who cannot discern the pattern. Correspondingly, there is some referential ambiguity.
    \item {\bf Current sentence}: Evidently, it's {\it his} own work; the sounds are modulated in ways undetectable to normal hearing, and even \underline{he} can't discern any pattern to them. \\
    \Circle Easy \Circle Moderate \APLcirc{\Circle} Hard \Circle Impossible\\
    {\bf Analysis}: Due to the referential ambiguity of the preceding pronoun "his", we do not know whose work it is, therefore it is hard to determine who "he" refers to. Correspondingly, there is a high level of referential ambiguity for the pronoun "he".
    \item {\bf Current sentence}: Evidently, it's {\it Reynold's} own work; the sounds are modulated in ways undetectable to normal hearing, and even \underline{James} can't discern any pattern to them. \\
    \Circle Easy \Circle Moderate \Circle Hard \APLcirc{\Circle} Impossible\\
    {\bf Analysis}: There are only two choices in the entity list: Leon Greco and Reynolds. The name "James" cannot refer to either of them, therefore entity disambiguation is impossible.
\end{enumerate}
\end{small}
}
\noindent Each worker $w$ was also instructed to rate the naturalness $nat(S, w)$ of the current sentence $S$, where a sentence was considered to sound natural if it was grammatical and not repetitive in terms of the words used for the entity mentions. The naturalness ratings ranged from 0: {\it Poor}, to 1: {\it Fair}, to 2: {\it Good}. \textcolor{black}{Below are examples illustrating ratings of naturalness of sentences which were provided as part of the instructions to the MTurk workers:
\begin{small}
\begin{enumerate}
    \item {\bf Current sentence}: \underline{Michael} never saw the need for them \underline{himself}, a little contact or anything that even sounded like contact would give \underline{him} more speed than \underline{he} could bear. \\
    \APLcirc{\Circle} Good \Circle Fair \Circle Poor \\
    {\bf Analysis}: The sentence is grammatical and appropriately uses the name "Michael" to ground the entity and personal pronouns for subsequent mentions.
    \item {\bf Current sentence}: \underline{Michael} never saw the need for them \underline{himself}, a little contact or anything that even sounded like contact would give \underline{Michael} more speed than \underline{he} could bear. \\
    \Circle Good \APLcirc{\Circle} Fair \Circle Poor \\
    {\bf Analysis}: The name "Michael" is unnecessarily used a second time, whereas a pronoun would have been a more natural choice. This makes the sentence sound a bit repetitive.
    \item {\bf Current sentence}: \underline{Michael} never saw the need for them \underline{himself}, a little contact or anything that even sounded like contact would give \underline{Michael} more speed than \underline{Michael} could bear. \\
    \Circle Good \Circle Fair \APLcirc{\Circle} Poor \\
    {\bf Analysis}: The name "Michael" is used three time, twice in a row, which makes the sentence sound very repetitive and thus unnatural.
\end{enumerate}
\end{small}
}

In total, 5 text fragments from the PoV dataset were evaluated by 5 workers each. The number of sentences and mentions in each paragraph are shown in Table \ref{tab:evaluation-statistics}. In total, a similar number of sentences was used for human evaluations in related tasks, such as \citep{kn:xu-etal-2012-paraphrasing,kn:clark-etal-2018-neural,kn:prabhumoye-etal-2018-style}. We also tested if the workers were consistent in their evaluations of the 3 versions, i.e. if previous sentences of any two versions are the same, ideally a worker should give the same ratings whenever the currently evaluated sentence is identical in these two versions.
Evaluations that were done at random or that were inconsistent were rejected and the tasks were resubmitted until each text had 5 evaluations.
%(more details below in Section~\ref{sec:human-results})
\begin{table}[ht]
\footnotesize
	\centering		
	\begin{tabular}{ c | c | c | c | c | c | c |}
		\cline{2-7} &
		\multicolumn{1}{c|}{$T_1$} & $T_2$ & $T_3$ & $T_4$ & $T_5$ & Total \\
		\cline{2-7}
		\hline
		\multicolumn{1}{|c|}{Sentences} & 10 & 8 & 11 & 12 & 10 & 51\\
		\hline
		\multicolumn{1}{|c|}{Mentions} & 21 & 16 & 26 & 30 & 50 & 143\\
		\hline
	\end{tabular}
	\caption{Number of sentences and mentions in each of the 5 text fragments $T_k$, and in total.}	
	\label{tab:evaluation-statistics}
\end{table}

\subsubsection{Human Evaluation Results}
\label{sec:human-results}

For each sentence $S$ evaluated by MTurk workers, we compute a {\it referential} score $ref(S)$ and a {\it naturalness} score $nat(S)$. Given the ambiguity $amb(m, w)$ assigned by worker $w$ to mention $m$, the referential score of sentence $S$ is calculated as follows:
\begin{equation}
%\small
%\resizebox{.89\linewidth}{!}{$
\!\!\!ref(S,w) = \frac{1}{|S|} \displaystyle\sum_{m \in S} amb(m,w) \cdot C(m,w)
\label{eq:ref-score}
\end{equation}
\begin{figure}[!t]
%\vspace*{-20pt}
\centering
	\includegraphics[width=0.6\textwidth]{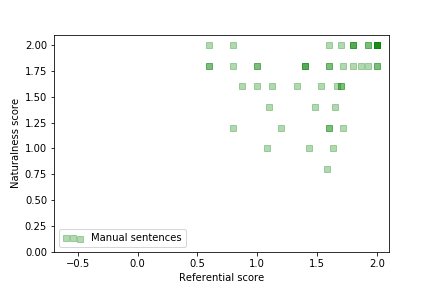}
	\caption{Referential and naturalness scores for each evaluated sentence in the Manual version.} \label{fig:all-samples-manual}
\end{figure}
\begin{figure}[!t]
%\vspace*{-20pt}
\centering
	\includegraphics[width=0.6\textwidth]{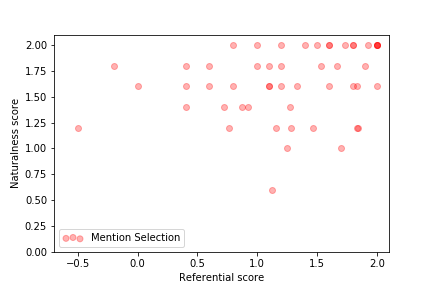}
	\caption{Referential and naturalness scores for each evaluated sentence in the Mention Selection version.} \label{fig:all-samples-gold}
\end{figure}
\begin{figure}[!t]
\centering
	\includegraphics[width=0.6\textwidth]{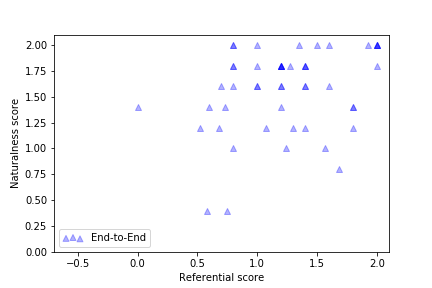}
	\caption{Referential and naturalness scores for each evaluated sentence in the End-to-End version.} \label{fig:all-samples-auto}
\end{figure}
\noindent where the correctness $C(m,w) = +1$ if $m$ is mapped by $w$ to the correct entity, and $C(m,w) = -1$ otherwise. If the mapping of $m$ was rated as impossible, the corresponding term in the sum above is set directly to $-2$. The referential score indicates how successful the mention string was in referring to the correct entity. If the worker mapped the mention to the correct entity, the easier the mapping, the higher the score. 
Conversely, if the mapping was incorrect, the more confident the worker was, the higher the penalty. Thus, the referential score ranges from $-2$ for a confident but incorrect entity mapping to $+2$ for a confident and correct mapping. A referential score of $-2$ is also given to an impossible mention-entity mapping caused by coreference errors. The overall referential and naturalness scores for a sentence are calculated by averaging over the 5 workers:
\begin{eqnarray}
%\small
ref(S) & \!\!=\!\! & \frac{1}{5}\sum_{w=1}^{5}ref(S,w)\\
nat(S) & \!\!=\!\! & \frac{1}{5}\sum_{w=1}^{5}nat(S,w)
%\label{eq:ref-nat-score}
\end{eqnarray}

Figures~\ref{fig:all-samples-manual}, \ref{fig:all-samples-gold} and \ref{fig:all-samples-auto} show scatter plots of the naturalness vs. referential scores for all sentences from the three versions. Darker colors indicate multiple sentences with the same scores. Sentences scoring high on both criteria should appear close to the top right corner, corresponding to the maximum possible scores $ref(S)=2$ and $nat(S) = 2$. In terms of naturalness, the distribution of the system generated sentences (Mention Selection and End-to-End) is comparable to the manually annotated annotations. We attribute this at least in part to the use of contextual BERT embedding, which helps in selecting mention strings that lead to syntactically correct sentences. The system has the tendency to select pronouns over noun phrases, which makes the generated sentences sound natural but may also introduce ambiguity, as evidenced in Figures~\ref{fig:all-samples-gold} and \ref{fig:all-samples-auto} where the system referential scores are distributed slighter farther away from the optimal.

The overall referential and naturalness scores of the generated mentions of the 3 versions are shown in Table~ \ref{table:human-evaluation-results}. If we map the $[-2, 2]$ score ranges to the $[0, 100]$ interval \textcolor{black}{so that 0 correspond to the minimum score possible}, in terms of referential scores the Mention Selection system obtains 81\% \textcolor{black}{of the possible maximum}, the End-to-End system obtains 80.75\%, compared to 87.25\% for our Manual annotations. In terms of naturalness, the Mention Selection system obtains 91\% \textcolor{black}{of the possible maximum}, the End-to-End system obtains 89.25\%, compared to 91.75\% for the Manual annotation.
%We also computed a {\it consistency} score for each worker by comparing their ratings on sentences that were the same in three versions.  
%Table~\ref{table:human-evaluation-results} reports consistency in terms of two measures: the root mean squared error (RMSE) and the mean absolute error (MAE) between the ratings of identical sentences in the two version, over all the workers.

\begin{table}[ht]
	\small
	\centering	
%	\begin{tabular}{| p{10mm} | p{6mm} |p{5mm} | p{5mm} | p{5mm}| p{8mm} | p{9mm}|}
	\begin{tabular}{|c| c|c|c|c| c| c|}
		\cline{2-7}
		\multicolumn{1}{c|}{} & \multicolumn{2}{c|}{End-to-End} & \multicolumn{2}{c|}{M. Selection} & \multicolumn{2}{c|}{Manual}\\
		\cline{2-7}
		\multicolumn{1}{c|}{} & $\mu$ & $\sigma$ & $\mu$ & $\sigma$ & $\mu$ & $\sigma$ \\
		\hline
		$ref$ & 1.23 & 0.46 & 1.24 & 0.59 & 1.49 & 0.42 \\
		\hline
		$nat$ & 1.57 & 0.40 & 1.64 & 0.34 & 1.67 & 0.32\\
       \hline
	\end{tabular}
	\caption{The mean $\mu$ and standard deviation $\sigma$ of the referential scores $ref$ and naturalness scores $nat$ over all sentences in the human evaluation.}
	\label{table:human-evaluation-results}
\end{table}

\section{Conclusion and Future Work}
\label{sec:conclusion}

We introduced the task of changing the narrative perspectives and proposed a pipeline architecture where a mention selection model is trained on data extracted from a corpus annotated with coreference chains.
We annotated a benchmark dataset with point of view changes, using text from a wide range of types of narratives.
Evaluations on a newly created PoV benchmark dataset showed that the proposed architecture substantially outperforms the baselines  and comes close to the manual annotation in terms of referential quality and naturalness. To account for the existence of multiple solutions to the PoV change of a text, we also ran and compared human evaluations of the system output and the manual annotations. To the best of our knowledge, the system presented in this paper is the first to address the problem of changing the narrative perspective of a text. To enable progress in this area, we are making the datasets and the code publicly available.
The source code and datasets are publicly available at \url{https://github.com/chenmike1986/change_pov}.

The approach introduced in this paper has the potential for a wide range of applications, such as the generation of self-diagnosis texts, or text written from a neutral, third-person perspective. The manipulation of narrative perspectives can also be valuable in fiction, where the aim would be to alter the reading experience.

In future work, we plan to extend the approach such that text between mentions is also allowed to change, which is expected to further improve the naturalness of the converted text.

% investigate the incorporation of attention and copy mechanisms in the LSTM models \textcolor{black}{to solve mention detection, coreference resolution, verb conjugation change and mention selection by one neural network system}.  We also plan to design neural architectures that are suitable for solving a more general approach to the task of changing the narrative perspective, where text between mentions is also allowed to change, to further improve the naturalness of the converted text. 
%  We currently do not consider the case when the number of the focus entity is plural, which is left for future work.

% Finally, we plan to design automatic evaluation metrics that take into account the existence of multiple good solutions for changing the point of view of any given input text.
%To further improve performance, we could maintain and update a vector representation of each entity in the text at each step in the LSTM models that are run from left to right, similar to \citet{kn:clark-etal-2018-neural}.

\section{Acknowledgements}
We would like to thank Dana Simionescu for the annotation of the PoV change dataset and Edmond Chang for providing connections to research on the use of point of view in literature and virtual worlds. In its initial stages, the project benefited from numerous discussions with Pooya Taherkhani. Finally, we would like to thank the anonymous reviewers for their detailed and constructive feedback.

\bibliography{clp20}

\begin{thebibliography}{79}
\expandafter\ifx\csname natexlab\endcsname\relax\def\natexlab#1{#1}\fi
\providecommand{\url}[1]{\texttt{#1}}
\providecommand{\href}[2]{#2}
\providecommand{\path}[1]{#1}
\providecommand{\DOIprefix}{doi:}
\providecommand{\ArXivprefix}{arXiv:}
\providecommand{\URLprefix}{URL: }
\providecommand{\Pubmedprefix}{pmid:}
\providecommand{\doi}[1]{\href{http://dx.doi.org/#1}{\path{#1}}}
\providecommand{\Pubmed}[1]{\href{pmid:#1}{\path{#1}}}
\providecommand{\bibinfo}[2]{#2}
\ifx\xfnm\relax \def\xfnm[#1]{\unskip,\space#1}\fi
%Type = Misc
\bibitem[{Asimov(1953)}]{kn:Issac}
\bibinfo{author}{Asimov, I.}, \bibinfo{year}{1953}.
\newblock \bibinfo{title}{Nobody here but --}.
\newblock
  \bibinfo{howpublished}{\url{https://www.goodreads.com/book/show/24967968-nobody-here-but}}.
\newblock \bibinfo{note}{Accessed October, 2020}.
%Type = Book
\bibitem[{Atwood(1998)}]{kn:Atwood98}
\bibinfo{author}{Atwood, M.}, \bibinfo{year}{1998}.
\newblock \bibinfo{title}{The Handmaid's Tale}.
\newblock \bibinfo{publisher}{Anchor Books}.
%Type = Inbook
\bibitem[{Belz et~al.(2010)Belz, Kow, Viethen and Gatt}]{belz:chapter10}
\bibinfo{author}{Belz, A.}, \bibinfo{author}{Kow, E.},
  \bibinfo{author}{Viethen, J.}, \bibinfo{author}{Gatt, A.},
  \bibinfo{year}{2010}.
\newblock \bibinfo{title}{Generating Referring Expressions in Context: The GREC
  Task Evaluation Challenges}. \bibinfo{publisher}{Springer Berlin Heidelberg},
  \bibinfo{address}{Berlin, Heidelberg}.
\newblock pp. \bibinfo{pages}{294--327}.
\newblock \URLprefix \url{https://doi.org/10.1007/978-3-642-15573-4_15},
  \DOIprefix\doi{10.1007/978-3-642-15573-4_15}.
%Type = Book
\bibitem[{Biss(2009)}]{kn:Biss09}
\bibinfo{author}{Biss, E.}, \bibinfo{year}{2009}.
\newblock \bibinfo{title}{Notes from No Man's Land}.
\newblock \bibinfo{publisher}{Graywolf Press}.
%Type = Article
\bibitem[{Black(2017)}]{black:gc17}
\bibinfo{author}{Black, D.}, \bibinfo{year}{2017}.
\newblock \bibinfo{title}{Why can i see my avatar? embodied visual engagement
  in the third-person video game}.
\newblock \bibinfo{journal}{Games and Culture} \bibinfo{volume}{12},
  \bibinfo{pages}{179--199}.
\newblock \URLprefix \url{https://doi.org/10.1177/1555412015589175},
  \DOIprefix\doi{10.1177/1555412015589175},
  \href{http://arxiv.org/abs/https://doi.org/10.1177/1555412015589175}{\tt
  arXiv:https://doi.org/10.1177/1555412015589175}.
%Type = Incollection
\bibitem[{Bohman(2000)}]{bohman:chapter18}
\bibinfo{author}{Bohman, J.}, \bibinfo{year}{2000}.
\newblock \bibinfo{title}{The importance of the second person: Interpretation,
  practical knowledge, and normative attitudes}, in:
  \bibinfo{booktitle}{Empathy And Agency: The Problem Of Understanding In The
  Human Sciences}.
\newblock \URLprefix \url{https://doi.org/10.4324/9780429500831}.
%Type = Article
\bibitem[{Bruny{\'{e}} et~al.(2009)Bruny{\'{e}}, Ditman, Mahoney, Augustyn and
  Taylor}]{Brunye:09}
\bibinfo{author}{Bruny{\'{e}}, T.T.}, \bibinfo{author}{Ditman, T.},
  \bibinfo{author}{Mahoney, C.R.}, \bibinfo{author}{Augustyn, J.S.},
  \bibinfo{author}{Taylor, H.A.}, \bibinfo{year}{2009}.
\newblock \bibinfo{title}{When {You} and {I} share perspectives: Pronouns
  modulate perspective taking during narrative comprehension}.
\newblock \bibinfo{journal}{Psychology Science} \bibinfo{volume}{20}.
%Type = Book
\bibitem[{Bryson(2010)}]{kn:Bryson10}
\bibinfo{author}{Bryson, B.}, \bibinfo{year}{2010}.
\newblock \bibinfo{title}{A Walk in the Woods}.
\newblock \bibinfo{publisher}{Broadway Books}.
%Type = Proceedings
\bibitem[{Campos et~al.(2021)Campos, Jorge, Jatowt, Bhatia and
  Finlayson}]{ws-2021-storytelling}
\bibinfo{editor}{Campos, R.}, \bibinfo{editor}{Jorge, A.M.},
  \bibinfo{editor}{Jatowt, A.}, \bibinfo{editor}{Bhatia, S.},
  \bibinfo{editor}{Finlayson, M.} (Eds.), \bibinfo{year}{2021}.
\newblock \bibinfo{title}{Proceedings of the Fourth International Workshop on
  Narrative Extraction from Texts (Text2Story)}, \bibinfo{address}{Lucca,
  Italy}.
\newblock \URLprefix \url{http://text2story21.inesctec.pt/}.
  \bibinfo{note}{{T}o appear}.
%Type = Misc
\bibitem[{Center(2018)}]{website:doctor}
\bibinfo{author}{Center, F.C.M.}, \bibinfo{year}{2018}.
\newblock \bibinfo{title}{Fulton county medical center main page}.
\newblock \URLprefix
  \url{http://fcmcpa.org/The-Importance-of-Healthy-Doctor-Patient-Relationships}.
  \bibinfo{note}{accessed October, 2018}.
%Type = Book
\bibitem[{Chiang(1991)}]{kn:Ted91}
\bibinfo{editor}{Chiang, T.} (Ed.), \bibinfo{year}{1991}.
\newblock \bibinfo{title}{Understand}.
\newblock \bibinfo{publisher}{Asimov's Science Fiction}.
%Type = Article
\bibitem[{Choifer(2018)}]{choifer:pp18}
\bibinfo{author}{Choifer, A.}, \bibinfo{year}{2018}.
\newblock \bibinfo{title}{A new understanding of the first-person and
  third-person perspectives}.
\newblock \bibinfo{journal}{Philosophical Papers} \bibinfo{volume}{47},
  \bibinfo{pages}{333--371}.
\newblock \URLprefix \url{https://doi.org/10.1080/05568641.2018.1450160},
  \DOIprefix\doi{10.1080/05568641.2018.1450160},
  \href{http://arxiv.org/abs/https://doi.org/10.1080/05568641.2018.1450160}{\tt
  arXiv:https://doi.org/10.1080/05568641.2018.1450160}.
%Type = Misc
\bibitem[{Chu(2013)}]{kn:Chu13}
\bibinfo{author}{Chu, J.}, \bibinfo{year}{2013}.
\newblock \bibinfo{title}{The water that falls on you from nowhere}.
\newblock
  \bibinfo{howpublished}{\url{https://www.tor.com/2013/02/20/the-water-that-falls-on-you-from-nowhere/}}.
\newblock \bibinfo{note}{Accessed October, 2020}.
%Type = Inproceedings
\bibitem[{Clark et~al.(2018)Clark, Ji and Smith}]{kn:clark-etal-2018-neural}
\bibinfo{author}{Clark, E.}, \bibinfo{author}{Ji, Y.}, \bibinfo{author}{Smith,
  N.A.}, \bibinfo{year}{2018}.
\newblock \bibinfo{title}{Neural text generation in stories using entity
  representations as context}, in: \bibinfo{booktitle}{Proceedings of the 2018
  Conference of the North {A}merican Chapter of the Association for
  Computational Linguistics: Human Language Technologies, Volume 1 (Long
  Papers)}, \bibinfo{publisher}{Association for Computational Linguistics},
  \bibinfo{address}{New Orleans, Louisiana}. pp. \bibinfo{pages}{2250--2260}.
\newblock \URLprefix \url{https://www.aclweb.org/anthology/N18-1204},
  \DOIprefix\doi{10.18653/v1/N18-1204}.
%Type = Article
\bibitem[{Collobert et~al.(2011)Collobert, Weston, Bottou, Karlen, Kavukcuoglu
  and Kuksa}]{JMLR:v12:collobert11a}
\bibinfo{author}{Collobert, R.}, \bibinfo{author}{Weston, J.},
  \bibinfo{author}{Bottou, L.}, \bibinfo{author}{Karlen, M.},
  \bibinfo{author}{Kavukcuoglu, K.}, \bibinfo{author}{Kuksa, P.},
  \bibinfo{year}{2011}.
\newblock \bibinfo{title}{Natural language processing (almost) from scratch}.
\newblock \bibinfo{journal}{Journal of Machine Learning Research}
  \bibinfo{volume}{12}, \bibinfo{pages}{2493--2537}.
\newblock \URLprefix \url{http://jmlr.org/papers/v12/collobert11a.html}.
%Type = Inproceedings
\bibitem[{Devlin et~al.(2019)Devlin, Chang, Lee and Toutanova}]{kn:Jacob18}
\bibinfo{author}{Devlin, J.}, \bibinfo{author}{Chang, M.W.},
  \bibinfo{author}{Lee, K.}, \bibinfo{author}{Toutanova, K.},
  \bibinfo{year}{2019}.
\newblock \bibinfo{title}{{BERT}: Pre-training of deep bidirectional
  transformers for language understanding}, in: \bibinfo{booktitle}{Proceedings
  of the 2019 Conference of the North {A}merican Chapter of the Association for
  Computational Linguistics: Human Language Technologies, Volume 1 (Long and
  Short Papers)}, \bibinfo{publisher}{Association for Computational
  Linguistics}, \bibinfo{address}{Minneapolis, Minnesota}. pp.
  \bibinfo{pages}{4171--4186}.
\newblock \URLprefix \url{https://www.aclweb.org/anthology/N19-1423},
  \DOIprefix\doi{10.18653/v1/N19-1423}.
%Type = Inproceedings
\bibitem[{Dozat and Manning(2017)}]{DBLP:conf/iclr/DozatM17}
\bibinfo{author}{Dozat, T.}, \bibinfo{author}{Manning, C.D.},
  \bibinfo{year}{2017}.
\newblock \bibinfo{title}{Deep biaffine attention for neural dependency
  parsing}, in: \bibinfo{booktitle}{5th International Conference on Learning
  Representations, {ICLR} 2017, Toulon, France, April 24-26, 2017, Conference
  Track Proceedings}, \bibinfo{publisher}{OpenReview.net}.
\newblock \URLprefix \url{https://openreview.net/forum?id=Hk95PK9le}.
%Type = Inproceedings
\bibitem[{Eisenberg and Finlayson(2016)}]{kn:eisenberg2016}
\bibinfo{author}{Eisenberg, J.D.}, \bibinfo{author}{Finlayson, M.A.},
  \bibinfo{year}{2016}.
\newblock \bibinfo{title}{Automatic identification of narrative diegesis and
  point of view}, in: \bibinfo{booktitle}{Proceedings of the 2nd Workshop on
  Computing News Storylines}.
\newblock \URLprefix \url{http://www.aclweb.org/anthology/W16-5705}.
%Type = Misc
\bibitem[{Evans(2015)}]{kn:Melanie}
\bibinfo{author}{Evans, M.T.}, \bibinfo{year}{2015}.
\newblock \bibinfo{title}{Narcissistic abuse - {Y}ou're damned if you do,
  damned if you don't}.
\newblock
  \bibinfo{howpublished}{\url{https://blog.melanietoniaevans.com/narcissistic-abuse-youre-damned-if-you-do-damned-if-you-dont/}}.
\newblock \bibinfo{note}{Accessed October, 2020}.
%Type = Inproceedings
\bibitem[{Ferreira et~al.(2018)Ferreira, Moussallem, K{\'a}d{\'a}r, Wubben and
  Krahmer}]{ferreira:acl18}
\bibinfo{author}{Ferreira, T.C.}, \bibinfo{author}{Moussallem, D.},
  \bibinfo{author}{K{\'a}d{\'a}r, {\'A}.}, \bibinfo{author}{Wubben, S.},
  \bibinfo{author}{Krahmer, E.}, \bibinfo{year}{2018}.
\newblock \bibinfo{title}{Neuralreg: An end-to-end approach to referring
  expression generation}, in: \bibinfo{booktitle}{Proceedings of the 56th
  Annual Meeting of the Association for Computational Linguistics (Volume 1:
  Long Papers)}, pp. \bibinfo{pages}{1959--1969}.
%Type = Article
\bibitem[{Fludernik(1993)}]{fludernik:aaa93}
\bibinfo{author}{Fludernik, M.}, \bibinfo{year}{1993}.
\newblock \bibinfo{title}{Second person fiction: Narrative "you" as addressee
  and/or protagonist}.
\newblock \bibinfo{journal}{AAA: Arbeiten aus Anglistik und Amerikanistik}
  \bibinfo{volume}{18}, \bibinfo{pages}{217--247}.
\newblock \URLprefix \url{http://www.jstor.org/stable/43023644}.
%Type = Book
\bibitem[{Flynn(2005)}]{kn:Flynn05}
\bibinfo{author}{Flynn, N.}, \bibinfo{year}{2005}.
\newblock \bibinfo{title}{Another Bullshit Night in Suck City}.
\newblock \bibinfo{publisher}{W.W. Norton}.
%Type = Inproceedings
\bibitem[{Friedrich and Pinkal(2015)}]{friedrich-pinkal-2015-discourse}
\bibinfo{author}{Friedrich, A.}, \bibinfo{author}{Pinkal, M.},
  \bibinfo{year}{2015}.
\newblock \bibinfo{title}{Discourse-sensitive automatic identification of
  generic expressions}, in: \bibinfo{booktitle}{Proceedings of the 53rd Annual
  Meeting of the Association for Computational Linguistics and the 7th
  International Joint Conference on Natural Language Processing (Volume 1: Long
  Papers)}, \bibinfo{publisher}{Association for Computational Linguistics},
  \bibinfo{address}{Beijing, China}. pp. \bibinfo{pages}{1272--1281}.
\newblock \URLprefix \url{https://www.aclweb.org/anthology/P15-1123},
  \DOIprefix\doi{10.3115/v1/P15-1123}.
%Type = Inproceedings
\bibitem[{Fu et~al.(2018)Fu, Tan, Peng, Zhao and Yan}]{kn:Fu17}
\bibinfo{author}{Fu, Z.}, \bibinfo{author}{Tan, X.}, \bibinfo{author}{Peng,
  N.}, \bibinfo{author}{Zhao, D.}, \bibinfo{author}{Yan, R.},
  \bibinfo{year}{2018}.
\newblock \bibinfo{title}{Style transfer in text: Exploration and evaluation},
  in: \bibinfo{booktitle}{Proceedings of {AAAI}}.
%Type = Inproceedings
\bibitem[{Gardent et~al.(2017)Gardent, Shimorina, Narayan and
  Perez-Beltrachini}]{gardent:acl17}
\bibinfo{author}{Gardent, C.}, \bibinfo{author}{Shimorina, A.},
  \bibinfo{author}{Narayan, S.}, \bibinfo{author}{Perez-Beltrachini, L.},
  \bibinfo{year}{2017}.
\newblock \bibinfo{title}{The {W}eb{NLG} challenge: Generating text from {RDF}
  data}, in: \bibinfo{booktitle}{Proceedings of the 10th International
  Conference on Natural Language Generation}, \bibinfo{publisher}{Association
  for Computational Linguistics}, \bibinfo{address}{Santiago de Compostela,
  Spain}. pp. \bibinfo{pages}{124--133}.
\newblock \URLprefix \url{https://www.aclweb.org/anthology/W17-3518},
  \DOIprefix\doi{10.18653/v1/W17-3518}.
%Type = Article
\bibitem[{Gast et~al.(2013)Gast, Tomozei and Le~Boudec}]{kn:Gast13}
\bibinfo{author}{Gast, N.G.}, \bibinfo{author}{Tomozei, D.C.},
  \bibinfo{author}{Le~Boudec, J.Y.}, \bibinfo{year}{2013}.
\newblock \bibinfo{title}{Optimal generation and storage scheduling in the
  presence of renewable forecast uncertainties} , \bibinfo{pages}{11}\URLprefix
  \url{http://infoscience.epfl.ch/record/183046}.
%Type = Article
\bibitem[{Grosz et~al.(1995)Grosz, Joshi and
  Weinstein}]{grosz-etal-1995-centering}
\bibinfo{author}{Grosz, B.J.}, \bibinfo{author}{Joshi, A.K.},
  \bibinfo{author}{Weinstein, S.}, \bibinfo{year}{1995}.
\newblock \bibinfo{title}{{C}entering: A framework for modeling the local
  coherence of discourse}.
\newblock \bibinfo{journal}{Computational Linguistics} \bibinfo{volume}{21},
  \bibinfo{pages}{203--225}.
\newblock \URLprefix \url{https://www.aclweb.org/anthology/J95-2003}.
%Type = Book
\bibitem[{Harrigan and Wardrip-Fruin(2007)}]{harrigan:book07}
\bibinfo{author}{Harrigan, P.}, \bibinfo{author}{Wardrip-Fruin, N.},
  \bibinfo{year}{2007}.
\newblock \bibinfo{title}{Second Person: Role-Playing and Story in Games and
  Playable Media}.
\newblock \bibinfo{publisher}{MIT Press}, \bibinfo{address}{Cambridge, MA}.
%Type = Article
\bibitem[{Hartung et~al.(2016)Hartung, Burke, Hagoort and Willems}]{Hartung:16}
\bibinfo{author}{Hartung, F.}, \bibinfo{author}{Burke, M.},
  \bibinfo{author}{Hagoort, P.}, \bibinfo{author}{Willems, R.M.},
  \bibinfo{year}{2016}.
\newblock \bibinfo{title}{Taking perspective: Personal pronouns affect
  experiential aspects of literary reading}.
\newblock \bibinfo{journal}{PLOS ONE} \bibinfo{volume}{11}.
%Type = Book
\bibitem[{Herr(1977)}]{kn:Michael77}
\bibinfo{editor}{Herr, M.} (Ed.), \bibinfo{year}{1977}.
\newblock \bibinfo{title}{Dispatches}.
\newblock \bibinfo{publisher}{Alfred A. Knopf}.
%Type = Techreport
\bibitem[{Heylighen and {M}arc Dewaele(1999)}]{kn:Heylighen99formalityof}
\bibinfo{author}{Heylighen, F.}, \bibinfo{author}{{M}arc Dewaele, J.},
  \bibinfo{year}{1999}.
\newblock \bibinfo{title}{Formality of Language: {D}efinition, measurement and
  behavioral determinants}.
\newblock \bibinfo{type}{Technical Report}.
%Type = Article
\bibitem[{Hochreiter and Schmidhuber(1997)}]{kn:Hochreiter97}
\bibinfo{author}{Hochreiter, S.}, \bibinfo{author}{Schmidhuber, J.},
  \bibinfo{year}{1997}.
\newblock \bibinfo{title}{Long short-term memory}.
\newblock \bibinfo{journal}{Neural Computation} \bibinfo{volume}{9},
  \bibinfo{pages}{1735--1780}.
\newblock \URLprefix \url{http://dx.doi.org/10.1162/neco.1997.9.8.1735},
  \DOIprefix\doi{10.1162/neco.1997.9.8.1735}.
%Type = Misc
\bibitem[{Honnibal and Montani(2017)}]{kn:Spacy}
\bibinfo{author}{Honnibal, M.}, \bibinfo{author}{Montani, I.},
  \bibinfo{year}{2017}.
\newblock \bibinfo{title}{Spacy}.
\newblock \bibinfo{howpublished}{\url{https://spacy.io/}}.
%Type = Inproceedings
\bibitem[{Hovy et~al.(2006)Hovy, Marcus, Palmer, Ramshaw and
  Weischedel}]{hovy-etal-2006-ontonotes}
\bibinfo{author}{Hovy, E.}, \bibinfo{author}{Marcus, M.},
  \bibinfo{author}{Palmer, M.}, \bibinfo{author}{Ramshaw, L.},
  \bibinfo{author}{Weischedel, R.}, \bibinfo{year}{2006}.
\newblock \bibinfo{title}{{O}nto{N}otes: The 90{\%} solution}, in:
  \bibinfo{booktitle}{Proceedings of the Human Language Technology Conference
  of the {NAACL}, Companion Volume: Short Papers},
  \bibinfo{publisher}{Association for Computational Linguistics},
  \bibinfo{address}{New York City, USA}. pp. \bibinfo{pages}{57--60}.
\newblock \URLprefix \url{https://www.aclweb.org/anthology/N06-2015}.
%Type = Article
\bibitem[{Jaeger(2010)}]{kn:Jaeger10}
\bibinfo{author}{Jaeger, T.F.}, \bibinfo{year}{2010}.
\newblock \bibinfo{title}{Redundancy and reduction: speakers manage syntactic
  information density.} , \bibinfo{pages}{23--62}\URLprefix
  \url{https://www.ncbi.nlm.nih.gov/pmc/articles/PMC2896231/}.
%Type = Book
\bibitem[{Jaeger and Buz(2016)}]{kn:Jaeger16}
\bibinfo{editor}{Jaeger, T.F.}, \bibinfo{editor}{Buz, E.} (Eds.),
  \bibinfo{year}{2016}.
\newblock \bibinfo{title}{Signal Reduction and Linguistic Encoding}.
\newblock \bibinfo{publisher}{John Wiley \& Sons, Inc.}
%Type = Inproceedings
\bibitem[{Jhamtani et~al.(2017)Jhamtani, Gangal, Hovy and
  Nyberg}]{kn:JhamtaniGHN17}
\bibinfo{author}{Jhamtani, H.}, \bibinfo{author}{Gangal, V.},
  \bibinfo{author}{Hovy, E.}, \bibinfo{author}{Nyberg, E.},
  \bibinfo{year}{2017}.
\newblock \bibinfo{title}{Shakespearizing modern language using copy-enriched
  sequence to sequence models}, in: \bibinfo{booktitle}{Proceedings of the
  Workshop on Stylistic Variation}, \bibinfo{publisher}{Association for
  Computational Linguistics}, \bibinfo{address}{Copenhagen, Denmark}. pp.
  \bibinfo{pages}{10--19}.
\newblock \URLprefix \url{https://www.aclweb.org/anthology/W17-4902},
  \DOIprefix\doi{10.18653/v1/W17-4902}.
%Type = Inproceedings
\bibitem[{Ji et~al.(2017)Ji, Tan, Martschat, Choi and
  Smith}]{kn:ji-etal-2017-dynamic}
\bibinfo{author}{Ji, Y.}, \bibinfo{author}{Tan, C.},
  \bibinfo{author}{Martschat, S.}, \bibinfo{author}{Choi, Y.},
  \bibinfo{author}{Smith, N.A.}, \bibinfo{year}{2017}.
\newblock \bibinfo{title}{Dynamic entity representations in neural language
  models}, in: \bibinfo{booktitle}{Proceedings of the 2017 Conference on
  Empirical Methods in Natural Language Processing},
  \bibinfo{publisher}{Association for Computational Linguistics},
  \bibinfo{address}{Copenhagen, Denmark}. pp. \bibinfo{pages}{1830--1839}.
\newblock \URLprefix \url{https://www.aclweb.org/anthology/D17-1195},
  \DOIprefix\doi{10.18653/v1/D17-1195}.
%Type = Article
\bibitem[{Jorge et~al.(2019)Jorge, Campos, Jatowt and
  Nunes}]{jorge_information_2019}
\bibinfo{author}{Jorge, A.M.}, \bibinfo{author}{Campos, R.},
  \bibinfo{author}{Jatowt, A.}, \bibinfo{author}{Nunes, S.},
  \bibinfo{year}{2019}.
\newblock \bibinfo{title}{Information {Processing} \& {Management} {Journal}
  {Special} {Issue} on {Narrative} {Extraction} from {Texts} ({Text2Story}):
  {Preface}}.
\newblock \bibinfo{journal}{Information Processing \& Management}
  \bibinfo{volume}{56}, \bibinfo{pages}{1771--1774}.
\newblock \URLprefix
  \url{https://www.sciencedirect.com/science/article/pii/S0306457319304455},
  \DOIprefix\doi{https://doi.org/10.1016/j.ipm.2019.05.004}.
%Type = Misc
\bibitem[{Key(2015)}]{kn:Harrison}
\bibinfo{author}{Key, H.S.}, \bibinfo{year}{2015}.
\newblock \bibinfo{title}{My dad tried to kill me with an alligator}.
\newblock
  \bibinfo{howpublished}{\url{https://www.outsideonline.com/1979226/my-dad-tried-kill-me-alligator}}.
\newblock \bibinfo{note}{Accessed October, 2020}.
%Type = Inproceedings
\bibitem[{Kingma and Ba(2015)}]{kn:Kingma15}
\bibinfo{author}{Kingma, D.P.}, \bibinfo{author}{Ba, J.L.},
  \bibinfo{year}{2015}.
\newblock \bibinfo{title}{Adam: A method for stochastic optimization}, in:
  \bibinfo{booktitle}{ICLR}.
%Type = Misc
\bibitem[{Koul(2016)}]{kn:Scaachi}
\bibinfo{author}{Koul, S.}, \bibinfo{year}{2016}.
\newblock \bibinfo{title}{There's no recipe for growing up}.
\newblock
  \bibinfo{howpublished}{\url{https://www.buzzfeed.com/scaachikoul/looking-for-my-mother-at-the-bottom-of-a-pot}}.
\newblock \bibinfo{note}{Accessed October, 2020}.
%Type = Book
\bibitem[{Kurosawa(1983)}]{kn:Kurosawa83}
\bibinfo{author}{Kurosawa, A.}, \bibinfo{year}{1983}.
\newblock \bibinfo{title}{Akira Kurosawa: Something Like an Autobiography}.
\newblock \bibinfo{publisher}{Vintage}.
%Type = Inproceedings
\bibitem[{Lee et~al.(2017)Lee, He, Lewis and Zettlemoyer}]{kn:Lee17}
\bibinfo{author}{Lee, K.}, \bibinfo{author}{He, L.}, \bibinfo{author}{Lewis,
  M.}, \bibinfo{author}{Zettlemoyer, L.}, \bibinfo{year}{2017}.
\newblock \bibinfo{title}{End-to-end neural coreference resolution}, in:
  \bibinfo{booktitle}{Proceedings of the 2017 Conference on Empirical Methods
  in Natural Language Processing}, \bibinfo{publisher}{Association for
  Computational Linguistics}, \bibinfo{address}{Copenhagen, Denmark}. pp.
  \bibinfo{pages}{188--197}.
\newblock \URLprefix \url{https://www.aclweb.org/anthology/D17-1018},
  \DOIprefix\doi{10.18653/v1/D17-1018}.
%Type = Article
\bibitem[{Lim and Reeves(2009)}]{lim:mp09}
\bibinfo{author}{Lim, S.}, \bibinfo{author}{Reeves, B.}, \bibinfo{year}{2009}.
\newblock \bibinfo{title}{Being in the game: Effects of avatar choice and point
  of view on psychophysiological responses during play}.
\newblock \bibinfo{journal}{Media Psychology} \bibinfo{volume}{12},
  \bibinfo{pages}{348--370}.
\newblock \URLprefix \url{https://doi.org/10.1080/15213260903287242},
  \DOIprefix\doi{10.1080/15213260903287242},
  \href{http://arxiv.org/abs/https://doi.org/10.1080/15213260903287242}{\tt
  arXiv:https://doi.org/10.1080/15213260903287242}.
%Type = Article
\bibitem[{van Lissa et~al.(2016)van Lissa, Caracciolo, van Duuren and van
  Leuveren}]{Lissa:16}
\bibinfo{author}{van Lissa, C.J.}, \bibinfo{author}{Caracciolo, M.},
  \bibinfo{author}{van Duuren, T.}, \bibinfo{author}{van Leuveren, B.},
  \bibinfo{year}{2016}.
\newblock \bibinfo{title}{Difficult empathy: The effect of narrative
  perspective on readers' engagement with a first-person narrator}.
\newblock \bibinfo{journal}{DIEGESIS} \bibinfo{volume}{5}.
%Type = Book
\bibitem[{Mani(2013)}]{kn:Mani13}
\bibinfo{author}{Mani, I.}, \bibinfo{year}{2013}.
\newblock \bibinfo{title}{Computational Modeling of Narrative}.
\newblock \bibinfo{publisher}{Morgan \& Claypool Publishers}.
%Type = Inproceedings
\bibitem[{Manning et~al.(2014)Manning, Surdeanu, Bauer, Finkel, Bethard and
  McClosky}]{manning-EtAl:2014:P14-5}
\bibinfo{author}{Manning, C.D.}, \bibinfo{author}{Surdeanu, M.},
  \bibinfo{author}{Bauer, J.}, \bibinfo{author}{Finkel, J.},
  \bibinfo{author}{Bethard, S.J.}, \bibinfo{author}{McClosky, D.},
  \bibinfo{year}{2014}.
\newblock \bibinfo{title}{The {Stanford} {CoreNLP} natural language processing
  toolkit}, in: \bibinfo{booktitle}{Association for Computational Linguistics
  (ACL) System Demonstrations}, pp. \bibinfo{pages}{55--60}.
\newblock \URLprefix \url{http://www.aclweb.org/anthology/P/P14/P14-5010}.
%Type = Misc
\bibitem[{Miller(2016)}]{kn:Mary}
\bibinfo{author}{Miller, M.}, \bibinfo{year}{2016}.
\newblock \bibinfo{title}{The 37}.
\newblock
  \bibinfo{howpublished}{\url{http://www.joylandmagazine.com/regions/south/37}}.
\newblock \bibinfo{note}{Accessed October, 2020}.
%Type = Proceedings
\bibitem[{Mitchell et~al.(2018)Mitchell, Huang, Ferraro and
  Misra}]{ws-2018-storytelling}
\bibinfo{editor}{Mitchell, M.}, \bibinfo{editor}{Huang, T.H.K.},
  \bibinfo{editor}{Ferraro, F.}, \bibinfo{editor}{Misra, I.} (Eds.),
  \bibinfo{year}{2018}.
\newblock \bibinfo{title}{Proceedings of the First Workshop on Storytelling},
  \bibinfo{publisher}{Association for Computational Linguistics},
  \bibinfo{address}{New Orleans, Louisiana}.
\newblock \URLprefix \url{https://www.aclweb.org/anthology/W18-1500},
  \DOIprefix\doi{10.18653/v1/W18-15}.
%Type = Article
\bibitem[{Modi et~al.(2017)Modi, Titov, Demberg, Sayeed and
  Pinkal}]{modi:tacl17}
\bibinfo{author}{Modi, A.}, \bibinfo{author}{Titov, I.},
  \bibinfo{author}{Demberg, V.}, \bibinfo{author}{Sayeed, A.},
  \bibinfo{author}{Pinkal, M.}, \bibinfo{year}{2017}.
\newblock \bibinfo{title}{Modeling semantic expectation: Using script knowledge
  for referent prediction}.
\newblock \bibinfo{journal}{Transactions of the Association for Computational
  Linguistics} \bibinfo{volume}{5}, \bibinfo{pages}{31--44}.
\newblock \URLprefix \url{https://www.aclweb.org/anthology/Q17-1003},
  \DOIprefix\doi{10.1162/tacl_a_00044}.
%Type = Book
\bibitem[{Mondimore(2006)}]{kn:Mondimore06}
\bibinfo{author}{Mondimore, F.M.}, \bibinfo{year}{2006}.
\newblock \bibinfo{title}{Bipolar Disorder: A Guide for Patients and Families}.
\newblock \bibinfo{publisher}{John Hopkins University Press}.
%Type = Misc
\bibitem[{Morrison(2015)}]{kn:Toni}
\bibinfo{author}{Morrison, T.}, \bibinfo{year}{2015}.
\newblock \bibinfo{title}{Sweetness}.
\newblock
  \bibinfo{howpublished}{\url{https://www.newyorker.com/magazine/2015/02/09/sweetness-2}}.
\newblock \bibinfo{note}{Accessed October, 2020}.
%Type = Inproceedings
\bibitem[{Newell and Cheung(2018)}]{newell-cheung-2018-constructing}
\bibinfo{author}{Newell, E.}, \bibinfo{author}{Cheung, J.C.},
  \bibinfo{year}{2018}.
\newblock \bibinfo{title}{Constructing a lexicon of relational nouns}, in:
  \bibinfo{booktitle}{Proceedings of the Eleventh International Conference on
  Language Resources and Evaluation ({LREC} 2018)},
  \bibinfo{publisher}{European Language Resources Association (ELRA)},
  \bibinfo{address}{Miyazaki, Japan}.
\newblock \URLprefix \url{https://www.aclweb.org/anthology/L18-1537}.
%Type = Inproceedings
\bibitem[{Pareti et~al.(2013)Pareti, O{'}Keefe, Konstas, Curran and
  Koprinska}]{pareti-etal-2013-automatically}
\bibinfo{author}{Pareti, S.}, \bibinfo{author}{O{'}Keefe, T.},
  \bibinfo{author}{Konstas, I.}, \bibinfo{author}{Curran, J.R.},
  \bibinfo{author}{Koprinska, I.}, \bibinfo{year}{2013}.
\newblock \bibinfo{title}{Automatically detecting and attributing indirect
  quotations}, in: \bibinfo{booktitle}{Proceedings of the 2013 Conference on
  Empirical Methods in Natural Language Processing},
  \bibinfo{publisher}{Association for Computational Linguistics},
  \bibinfo{address}{Seattle, Washington, USA}. pp. \bibinfo{pages}{989--999}.
\newblock \URLprefix \url{https://www.aclweb.org/anthology/D13-1101}.
%Type = Article
\bibitem[{Pauen(2012)}]{pauen:inquiry12}
\bibinfo{author}{Pauen, M.}, \bibinfo{year}{2012}.
\newblock \bibinfo{title}{The second-person perspective}.
\newblock \bibinfo{journal}{Inquiry} \bibinfo{volume}{55},
  \bibinfo{pages}{33--49}.
\newblock \URLprefix \url{https://doi.org/10.1080/0020174X.2012.643623},
  \DOIprefix\doi{10.1080/0020174X.2012.643623},
  \href{http://arxiv.org/abs/https://doi.org/10.1080/0020174X.2012.643623}{\tt
  arXiv:https://doi.org/10.1080/0020174X.2012.643623}.
%Type = Book
\bibitem[{van Peer and Chatman(2001)}]{kn:Willie01}
\bibinfo{editor}{van Peer, W.}, \bibinfo{editor}{Chatman, S.} (Eds.),
  \bibinfo{year}{2001}.
\newblock \bibinfo{title}{New Perspectives on Narrative Perspective}.
\newblock The Margins of Literature, \bibinfo{publisher}{SUNY Press}.
%Type = Misc
\bibitem[{Platform(2018)}]{website:mother}
\bibinfo{author}{Platform, B.E.}, \bibinfo{year}{2018}.
\newblock \bibinfo{title}{Parenting advice article}.
\newblock \URLprefix
  \url{https://www.betterhelp.com/advice/parenting/what-is-a-toxic-mother-and-how-does-she-affect-relationships/}.
  \bibinfo{note}{accessed October, 2018}.
%Type = Article
\bibitem[{Poesio et~al.(2004)Poesio, Stevenson, Di~Eugenio and
  Hitzeman}]{poesi:cl04}
\bibinfo{author}{Poesio, M.}, \bibinfo{author}{Stevenson, R.},
  \bibinfo{author}{Di~Eugenio, B.}, \bibinfo{author}{Hitzeman, J.},
  \bibinfo{year}{2004}.
\newblock \bibinfo{title}{{C}entering: A parametric theory and its
  instantiations}.
\newblock \bibinfo{journal}{Computational Linguistics} \bibinfo{volume}{30},
  \bibinfo{pages}{309--363}.
\newblock \URLprefix \url{https://www.aclweb.org/anthology/J04-3003},
  \DOIprefix\doi{10.1162/0891201041850911}.
%Type = Book
\bibitem[{Pollan(2018)}]{kn:Pollan18}
\bibinfo{author}{Pollan, M.}, \bibinfo{year}{2018}.
\newblock \bibinfo{title}{How to Change Your Mind}.
\newblock \bibinfo{publisher}{Penguin Press}.
%Type = Misc
\bibitem[{Possinger(2018)}]{website:lawyer}
\bibinfo{author}{Possinger, J.}, \bibinfo{year}{2018}.
\newblock \bibinfo{title}{5 signs of a good lawyer}.
\newblock \URLprefix
  \url{https://medium.com/possingerlaw/5-signs-of-a-good-lawyer-b846b6e892e7}.
  \bibinfo{note}{accessed October, 2020}.
%Type = Inproceedings
\bibitem[{Prabhumoye et~al.(2018)Prabhumoye, Tsvetkov, Salakhutdinov and
  Black}]{kn:prabhumoye-etal-2018-style}
\bibinfo{author}{Prabhumoye, S.}, \bibinfo{author}{Tsvetkov, Y.},
  \bibinfo{author}{Salakhutdinov, R.}, \bibinfo{author}{Black, A.W.},
  \bibinfo{year}{2018}.
\newblock \bibinfo{title}{Style transfer through back-translation}, in:
  \bibinfo{booktitle}{Proceedings of the 56th Annual Meeting of the Association
  for Computational Linguistics (Volume 1: Long Papers)},
  \bibinfo{publisher}{Association for Computational Linguistics},
  \bibinfo{address}{Melbourne, Australia}. pp. \bibinfo{pages}{866--876}.
\newblock \URLprefix \url{https://www.aclweb.org/anthology/P18-1080},
  \DOIprefix\doi{10.18653/v1/P18-1080}.
%Type = Inproceedings
\bibitem[{Pradhan et~al.(2012)Pradhan, Moschitti, Xue, Uryupina and
  Zhang}]{kn:pradhan-etal-conll-st-2012-ontonotes}
\bibinfo{author}{Pradhan, S.}, \bibinfo{author}{Moschitti, A.},
  \bibinfo{author}{Xue, N.}, \bibinfo{author}{Uryupina, O.},
  \bibinfo{author}{Zhang, Y.}, \bibinfo{year}{2012}.
\newblock \bibinfo{title}{{CoNLL-2012} shared task: Modeling multilingual
  unrestricted coreference in {OntoNotes}}, in:
  \bibinfo{booktitle}{{Proceedings of the Sixteenth Conference on Computational
  Natural Language Learning (CoNLL 2012)}}, \bibinfo{address}{Jeju, Korea}.
%Type = Incollection
\bibitem[{Prince(1981)}]{prince:chapter81}
\bibinfo{author}{Prince, E.F.}, \bibinfo{year}{1981}.
\newblock \bibinfo{title}{Toward a taxonomy of given-new information}, in:
  \bibinfo{editor}{Cole, P.} (Ed.), \bibinfo{booktitle}{Syntax and semantics:
  Vol. 14. Radical Pragmatics}. \bibinfo{publisher}{Academic Press},
  \bibinfo{address}{New York}, pp. \bibinfo{pages}{223--255}.
%Type = Inproceedings
\bibitem[{Rao and Tetreault(2018)}]{kn:rao-tetreault-2018-dear}
\bibinfo{author}{Rao, S.}, \bibinfo{author}{Tetreault, J.},
  \bibinfo{year}{2018}.
\newblock \bibinfo{title}{Dear sir or madam, may {I} introduce the {GYAFC}
  dataset: Corpus, benchmarks and metrics for formality style transfer}, in:
  \bibinfo{booktitle}{Proceedings of the 2018 Conference of the North
  {A}merican Chapter of the Association for Computational Linguistics: Human
  Language Technologies, Volume 1 (Long Papers)},
  \bibinfo{publisher}{Association for Computational Linguistics},
  \bibinfo{address}{New Orleans, Louisiana}. pp. \bibinfo{pages}{129--140}.
\newblock \URLprefix \url{https://www.aclweb.org/anthology/N18-1012},
  \DOIprefix\doi{10.18653/v1/N18-1012}.
%Type = Article
\bibitem[{Salem et~al.(2017)Salem, Weskott and Holler}]{Salem:17}
\bibinfo{author}{Salem, S.}, \bibinfo{author}{Weskott, T.},
  \bibinfo{author}{Holler, A.}, \bibinfo{year}{2017}.
\newblock \bibinfo{title}{Does narrative perspective influence readers'
  perspective-taking? {An} empirical study on free indirect discourse,
  psycho-narration and first-person narration}.
\newblock \bibinfo{journal}{Glossa: A Journal of General Linguistics}
  \bibinfo{volume}{2}.
%Type = Misc
\bibitem[{Samatar(2013)}]{kn:Sofia}
\bibinfo{author}{Samatar, S.}, \bibinfo{year}{2013}.
\newblock \bibinfo{title}{Selkie stories are for losers}.
\newblock
  \bibinfo{howpublished}{\url{http://strangehorizons.com/fiction/selkie-stories-are-for-losers/}}.
\newblock \bibinfo{note}{Accessed October, 2020}.
%Type = Book
\bibitem[{Sartre(1969)}]{kn:Sartre69}
\bibinfo{author}{Sartre, J.P.}, \bibinfo{year}{1969}.
\newblock \bibinfo{title}{Nausea}.
\newblock \bibinfo{publisher}{New Directions}.
%Type = Incollection
\bibitem[{Shen et~al.(2017)Shen, Lei, Barzilay and Jaakkola}]{kn:Shen17}
\bibinfo{author}{Shen, T.}, \bibinfo{author}{Lei, T.},
  \bibinfo{author}{Barzilay, R.}, \bibinfo{author}{Jaakkola, T.},
  \bibinfo{year}{2017}.
\newblock \bibinfo{title}{Style transfer from non-parallel text by
  cross-alignment}, in: \bibinfo{editor}{Guyon, I.}, \bibinfo{editor}{Luxburg,
  U.V.}, \bibinfo{editor}{Bengio, S.}, \bibinfo{editor}{Wallach, H.},
  \bibinfo{editor}{Fergus, R.}, \bibinfo{editor}{Vishwanathan, S.},
  \bibinfo{editor}{Garnett, R.} (Eds.), \bibinfo{booktitle}{Advances in Neural
  Information Processing Systems 30}. \bibinfo{publisher}{Curran Associates,
  Inc.}, pp. \bibinfo{pages}{6830--6841}.
\newblock \URLprefix
  \url{http://papers.nips.cc/paper/7259-style-transfer-from-non-parallel-text-by-cross-alignment.pdf}.
%Type = Inbook
\bibitem[{Smith(2003)}]{smith_2003}
\bibinfo{author}{Smith, C.S.}, \bibinfo{year}{2003}.
\newblock \bibinfo{title}{Introduction}. \bibinfo{publisher}{Cambridge
  University Press}.
\newblock Cambridge Studies in Linguistics, p. \bibinfo{pages}{1–4}.
\newblock \DOIprefix\doi{10.1017/CBO9780511615108.002}.
%Type = Book
\bibitem[{Solnit(2013)}]{kn:Solnit13}
\bibinfo{author}{Solnit, R.}, \bibinfo{year}{2013}.
\newblock \bibinfo{title}{The Faraway Nearby}.
\newblock \bibinfo{publisher}{Viking}.
%Type = Article
\bibitem[{Strube and Hahn(1999)}]{strube:cl99}
\bibinfo{author}{Strube, M.}, \bibinfo{author}{Hahn, U.}, \bibinfo{year}{1999}.
\newblock \bibinfo{title}{Functional centering: Grounding referential coherence
  in information structure}.
\newblock \bibinfo{journal}{Computational Linguistics} \bibinfo{volume}{25},
  \bibinfo{pages}{309–344}.
%Type = Techreport
\bibitem[{Underwood(2009)}]{kn:performative}
\bibinfo{author}{Underwood, W.}, \bibinfo{year}{2009}.
\newblock \bibinfo{title}{Examples of Performative Sentences in Presidential
  Records}.
\newblock \bibinfo{type}{Technical Report}.
%Type = Misc
\bibitem[{Ungerleider(2016)}]{kn:Shoshana}
\bibinfo{author}{Ungerleider, S.}, \bibinfo{year}{2016}.
\newblock \bibinfo{title}{I'm a doctor. {P}reparing you for death is as much a
  part of my job as saving lives}.
\newblock
  \bibinfo{howpublished}{\url{https://www.vox.com/2015/10/19/9554583/doctor-good-death}}.
\newblock \bibinfo{note}{Accessed October, 2020}.
%Type = Book
\bibitem[{Vonnegut(1999)}]{kn:Vonnegut99}
\bibinfo{author}{Vonnegut, K.}, \bibinfo{year}{1999}.
\newblock \bibinfo{title}{Mother Night}.
\newblock \bibinfo{publisher}{Dial Press Trade Paperback}.
%Type = Inproceedings
\bibitem[{Walker(1993)}]{walker:centering93}
\bibinfo{author}{Walker, M.A.}, \bibinfo{year}{1993}.
\newblock \bibinfo{title}{Initial contexts and shifting centers}, in:
  \bibinfo{booktitle}{Proceedings of the Workshop on Centering},
  \bibinfo{address}{University of Pennsylvania, Philadelphia}.
%Type = Incollection
\bibitem[{Weischedel et~al.(2011)Weischedel, Pradhan, Ramshaw
  et~al.}]{weischedel2011ontonotes}
\bibinfo{author}{Weischedel, R.}, \bibinfo{author}{Pradhan, S.},
  \bibinfo{author}{Ramshaw, L.}, et~al., \bibinfo{year}{2011}.
\newblock \bibinfo{title}{Ontonotes release 4.0}, in:
  \bibinfo{booktitle}{Handbook of Natural Language Processing and Machine
  Translation: DARPA Global Autonomous Language Exploitation}.
  \bibinfo{publisher}{Springer}.
%Type = Article
\bibitem[{Wiebe(1994)}]{kn:Wiebe94}
\bibinfo{author}{Wiebe, J.M.}, \bibinfo{year}{1994}.
\newblock \bibinfo{title}{Tracking point of view in narrative}.
\newblock \bibinfo{journal}{Computational Linguistics} \bibinfo{volume}{20},
  \bibinfo{pages}{233--287}.
\newblock \URLprefix \url{http://www.aclweb.org/anthology/J94-2004}.
%Type = Inproceedings
\bibitem[{Xu et~al.(2012)Xu, Ritter, Dolan, Grishman and
  Cherry}]{kn:xu-etal-2012-paraphrasing}
\bibinfo{author}{Xu, W.}, \bibinfo{author}{Ritter, A.}, \bibinfo{author}{Dolan,
  B.}, \bibinfo{author}{Grishman, R.}, \bibinfo{author}{Cherry, C.},
  \bibinfo{year}{2012}.
\newblock \bibinfo{title}{Paraphrasing for style}, in:
  \bibinfo{booktitle}{Proceedings of {COLING} 2012}, \bibinfo{publisher}{The
  COLING 2012 Organizing Committee}, \bibinfo{address}{Mumbai, India}. pp.
  \bibinfo{pages}{2899--2914}.
\newblock \URLprefix \url{https://www.aclweb.org/anthology/C12-1177}.

\end{thebibliography}

\end{document}